\documentclass[letterpaper, 10 pt, conference]{ieeeconf}  

\IEEEoverridecommandlockouts                              

\usepackage[dvipsnames]{xcolor} 
\usepackage{tikz} 
\usetikzlibrary{calc, arrows.meta, positioning} 
\usepackage{multirow} 
\usepackage{algorithm} 
\usepackage{algpseudocode} 
\usepackage{array} 
\usepackage{booktabs} 
\usepackage{amssymb}  
\usepackage{mathtools, empheq,eqparbox}  
\usepackage{subcaption} 
\usepackage{graphicx} 
\usepackage{makecell,balance}
\usepackage{float} 
\usepackage[most]{tcolorbox} 
\usepackage{svg} 
\usepackage{pdfpages} 
\usepackage[unicode=true,hyperfootnotes=false]{hyperref} 

\usepackage{subcaption}

\usepackage[letterpaper]{geometry}
\DeclareCaptionLabelSeparator{periodspace}{.\quad}
\usepackage[binary-units=true]{siunitx}

\usepackage{pifont}
\newcommand*\colourcheck[1]{%
  \expandafter\newcommand\csname #1check\endcsname{\textcolor{#1}{\ding{52}}}%
}
\colourcheck{blue}
\colourcheck{green}
\colourcheck{red}

\newcommand*\colourcross[1]{%
  \expandafter\newcommand\csname #1cross\endcsname{\textcolor{#1}{\ding{55}}}%
}
\colourcross{blue}
\colourcross{green}
\colourcross{red}

\usepackage[dvipsnames]{xcolor}

\title{\LARGE \bf
Robust MADER: Decentralized and Asynchronous Multiagent Trajectory Planner Robust to Communication Delay
}
 
\newcommand{\trajANew}{traj\textsubscript{A\textsubscript{new}}}
\newcommand{\trajA}{traj\textsubscript{A}}
\newcommand{\trajAPrev}{ traj\textsubscript{A\textsubscript{prev}} }

\newcommand{\trajBNew}{traj\textsubscript{B\textsubscript{new}}}
\newcommand{\trajB}{traj\textsubscript{B}}
\newcommand{\trajBPrev}{ traj\textsubscript{B\textsubscript{prev}} }

\newcommand{\trajJNew}{traj\textsubscript{J\textsubscript{new}}}
\newcommand{\trajJ}{traj\textsubscript{J}}
\newcommand{\trajJPrev}{ traj\textsubscript{J\textsubscript{prev}} }

\newcommand{\delayParameter}{\ensuremath{\delta_\text{DC}}}
\newcommand{\delayActual}{\ensuremath{\delta_\text{actual}}}
\newcommand{\delayActualMax}{\ensuremath{\delta_\text{max}}}
\newcommand{\delayIntroduced}{\ensuremath{\delta_\text{introd}}}

\newcommand{\AgentA}{{\color{red}Agent A}}

\newcommand{\AgentB}{{\color{blue}Agent B}}

\newcommand{\NeccessaryCond}{\ensuremath{\delayParameter\ge\delayActualMax{}}}
\newcommand{\NotNeccessaryCond}{\ensuremath{\delayActualMax{}\ge\delayParameter}}

\definecolor{opt_color}{RGB}{203,255,182}
\definecolor{check_color}{RGB}{195,218,255}
\definecolor{recheck_color}{RGB}{255,176,176}
\definecolor{delaycheck_color}{RGB}{255,228,181}

\newcommand\mybox[2][]{\tikz[overlay]\node[fill=blue!20,inner sep=0.6pt, anchor=text, rectangle, rounded corners=0mm,#1] {#2};\phantom{#2}}

\newcommand{\OptimizationStep}{\mybox[fill=opt_color]{\bf{Optimization}}}
\newcommand{\CheckStep}{\mybox[fill=check_color]{\bf{Check}}}
\newcommand{\RecheckStep}{\mybox[fill=recheck_color]{\bf{Recheck}}}
\newcommand{\DelayCheckStep}{\mybox[fill=delaycheck_color]{\bf{Delay Check}}}

\newcommand{\OStep}{\mybox[fill=opt_color]{\bf{O}}}
\newcommand{\CStep}{\mybox[fill=check_color]{\bf{C}}}
\newcommand{\RStep}{\mybox[fill=recheck_color]{\bf{R}}}
\newcommand{\DCStep}{\mybox[fill=delaycheck_color]{\bf{DC}}} 

\newcommand{\OStepA}{\mybox[fill=opt_color]{\bf{O}\textsubscript{A}}}
\newcommand{\CStepA}{\mybox[fill=check_color]{\bf{C}\textsubscript{A}}}
\newcommand{\RStepA}{\mybox[fill=recheck_color]{\bf{R}\textsubscript{A}}}
\newcommand{\DCStepA}{\mybox[fill=delaycheck_color]{\bf{DC}\textsubscript{A}}} 

\newcommand{\OStepB}{\mybox[fill=opt_color]{\bf{O}\textsubscript{B}}}
\newcommand{\CStepB}{\mybox[fill=check_color]{\bf{C}\textsubscript{B}}}
\newcommand{\RStepB}{\mybox[fill=recheck_color]{\bf{R}\textsubscript{B}}}
\newcommand{\DCStepB}{\mybox[fill=delaycheck_color]{\bf{DC}\textsubscript{B}}}

\newcommand{\intaccelsquared}{\ensuremath{\int\left\Vert\mathbf{a}\right\Vert^2dt}}
\newcommand{\intjerksquared}{\ensuremath{\int\left\Vert\mathbf{j}\right\Vert^2dt}}

\newcommand{\MADER}{\textbf{MADER}}
\newcommand{\RMADER}{\textbf{RMADER}}
\newcommand{\EGOswarm}{\textbf{EGO-Swarm}}
\newcommand{\SlowEGOswarm}{\textbf{Slow EGO-Swarm}}

\author{Kota Kondo, Jesus Tordesillas, Reinaldo Figueroa, \\ Juan Rached, Joseph Merkel, Parker C. Lusk, and Jonathan P. How
\thanks{Aerospace Controls Laboratory, MIT, 77 Massachusetts Ave, Cambridge, MA, USA
{\tt \{kkondo, jtorde, reyfp, jrached, jamerkel, plusk, jhow\}@mit.edu}}}%

\begin{document}

\maketitle
\thispagestyle{plain}
\pagestyle{plain}


\begin{abstract}

Although communication delays can disrupt multiagent systems, most of the existing multiagent trajectory planners lack a strategy to address this issue. 
State-of-the-art approaches typically assume perfect communication environments, which is hardly realistic in real-world experiments. 
This paper presents Robust MADER (RMADER), a decentralized and asynchronous multiagent trajectory planner that can handle communication delays among agents. 
By broadcasting both the newly optimized trajectory and the committed trajectory, and by performing a delay check step, RMADER is able to guarantee safety even under communication delay.
RMADER was validated through extensive simulation and hardware flight experiments and achieved a 100\% success rate of collision-free trajectory generation, outperforming state-of-the-art approaches. \\

\end{abstract}





\section*{Supplementary Material}
\textbf{Video}: \href{https://youtu.be/vH09kwJOBYs}{https://youtu.be/vH09kwJOBYs} \\


\section{INTRODUCTION}\label{sec:intro}

Multiagent UAV trajectory planning has been extensively studied in the literature for its wide range of applications. These planners can be centralized \cite{park_efficient_2020, sharon_conflict-based_2015, robinson_efficient_2018} (one machine plans every agent's trajectory) or decentralized \cite{tordesillas_mader_2022, zhou_ego-swarm_2020, lusk_distributed_2020} (each agent plans its own trajectory).
Decentralized planners are more scalable and robust to failures of the centralized machine. Despite these advantages, a decentralized scheme requires communication between the agents, and communication delays could potentially introduce failure in the trajectory deconfliction between the agents, which is essential to guarantee safety~\cite{gielis_critical_2022}. 
multiagent planners can also be classified according to whether or not they are asynchronous. In an asynchronous setting, each agent independently triggers the planning step without considering the planning status of other agents. Asynchronous approaches do not require a synchronous mechanism among agents and therefore more scalable than synchronous approaches, but they are also more susceptible to communication delays since agents are planning and executing trajectories independently.

\begin{figure}[!t]
    \centering
    \subfloat[Hardware experiments: 6 UAVs running RMADER onboard. Despite the existence of communication delays between the agents, RMADER can guarantee safety. \label{fig:mader_hw_6_uavs}]{\includegraphics[width=0.9\columnwidth]{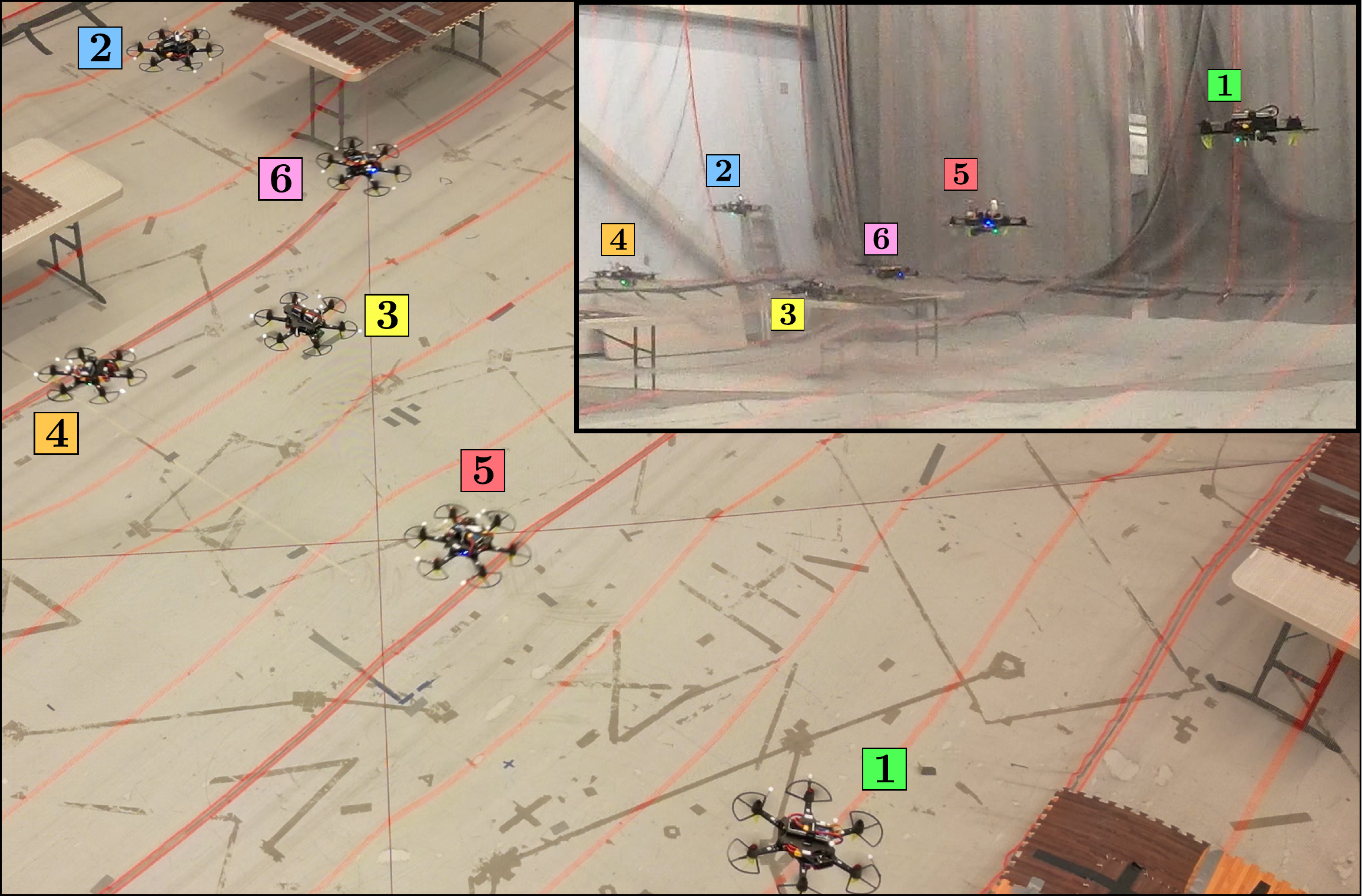}}\\
    
    \subfloat[Simulation experiments: 50 UAVs in a circle configuration successfully exchange their positions despite the \textbf{\SI{100}{\ms}} communication delay introduced on purpose. The color denotes the velocity (red higher and blue lower).]{\includegraphics[width=0.9\columnwidth]{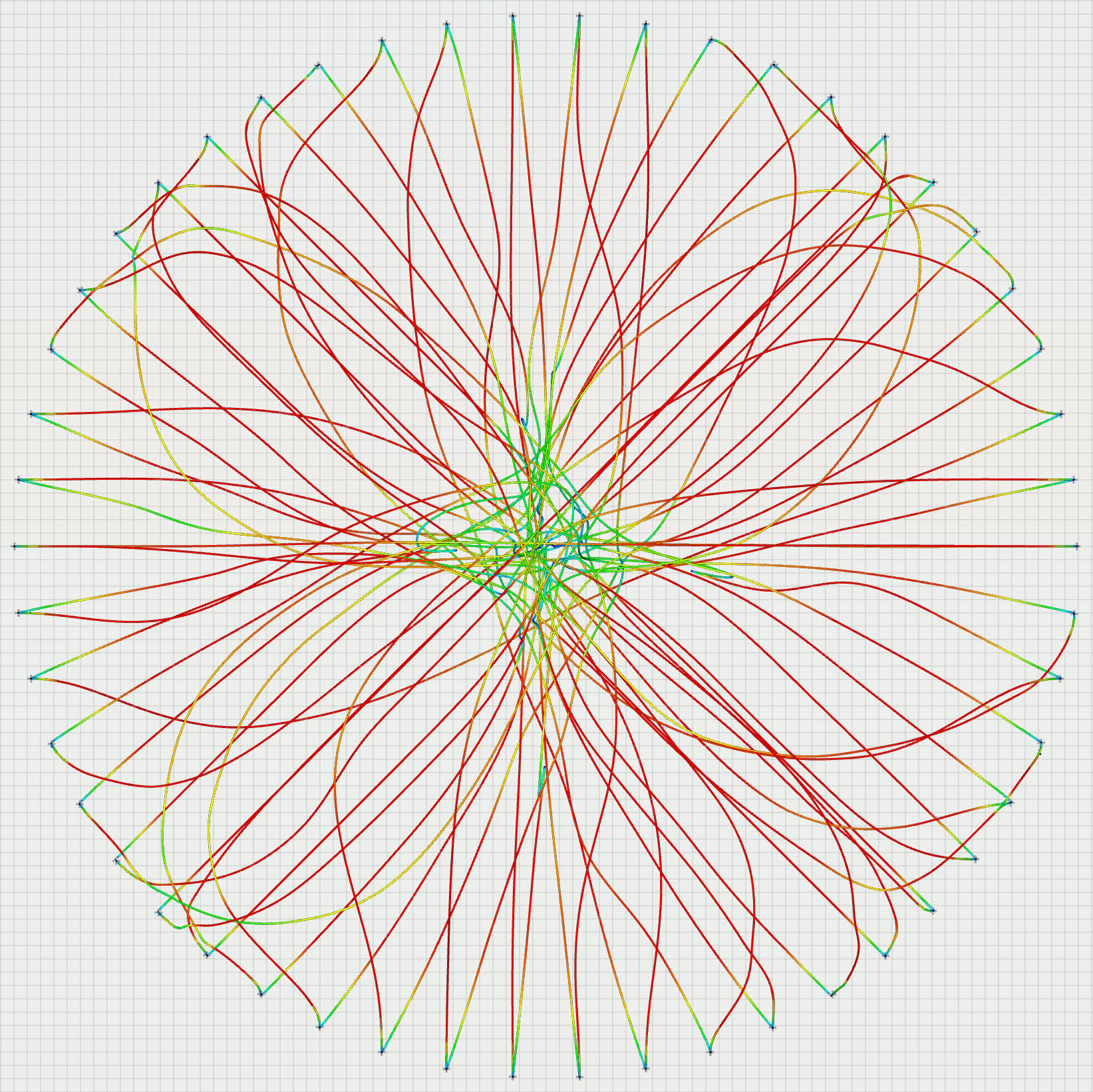}}
    \setlength{\belowcaptionskip}{-2em}
    \caption{\centering RMADER Hardware and Simulation Results}
    \vspace{-0.25em}
\end{figure}

Many decentralized state-of-the-art trajectory planners do not consider communication delays or explicitly state assumptions about communication. 
For example, the planners presented in \textbf{SCP}~\cite{chen_decoupled_2015}, \textbf{decNS}~\cite{liu_towards_2018}, and \textbf{LSC}~\cite{park_online_2022} are decentralized and synchronous, but SCP and decNS implicitly and LSC explicitly assume a perfect communication environment without any communication delays.

\newcommand{\NoRed}{\textbf{\textcolor{red}{No}}}
\newcommand{\YesGreen}{\textbf{\textcolor{ForestGreen}{Yes}}}

\begin{table}[!t]
    \renewcommand{\arraystretch}{1.4}
    \scriptsize
    \begin{centering}
    \caption{\centering State-of-the-art Decentralized Multiagent Planners.}
    \label{tab:state_of_the_art_comparison}
    \begin{tabular}{>{\centering\arraybackslash}m{0.2\columnwidth} >{\centering\arraybackslash}m{0.2\columnwidth} >{\centering\arraybackslash}m{0.2\columnwidth} >{\centering\arraybackslash}m{0.2\columnwidth}}
    \toprule 
    \textbf{Method} & \textbf{Asynchronous?} & \textbf{Handles Comm. Delay?} & \textbf{\makecell{Hardware \\ Demonstration}}\tabularnewline
    \hline 
    \hline 
    \textbf{SCP~\cite{chen_decoupled_2015} decNS~\cite{liu_towards_2018} LSC~\cite{park_online_2022}} & \NoRed{} & \NoRed{} & \YesGreen{} \tabularnewline
    \hline 
    \textbf{decMPC~\cite{toumieh_decentralized_2022}} & \NoRed{} & \YesGreen{} & \NoRed{} \tabularnewline
    \hline
    \textbf{decGroup~\cite{hou_enhanced_2022}} & \YesGreen{}/\NoRed{}\footnotemark[2] & \NoRed{} & \YesGreen{} \tabularnewline
    \hline 
    \textbf{ADPP~\cite{ cap_asynchronous_2013}} & \YesGreen{}\footnotemark[3] & \NoRed{} & \YesGreen{} \tabularnewline
    \hline
    \textbf{MADER~\cite{tordesillas_mader_2022}} & \YesGreen{} & \NoRed{} & \NoRed{} \tabularnewline
    \hline 
    \textbf{EGO-Swarm~\cite{zhou_ego-swarm_2020}} & \YesGreen{} & \NoRed{} & \YesGreen{} \tabularnewline
    \hline 
    \textbf{AsyncBVC~\cite{senbaslar_asynchronous_2022}} & \YesGreen{} & \YesGreen{}  & \NoRed{} \tabularnewline
    \hline
    \textbf{RMADER \ (proposed)} & \YesGreen{} & \YesGreen{} & \YesGreen{} \tabularnewline
    \bottomrule
    \end{tabular}
    \par\end{centering}
\vspace*{0.5em}
\footnotesize{$^2$ \!\!\!\! decGroup triggers joint-optimization in dense environments and switches to a centralized, synchronous planner.} \\
\footnotesize{$^3$  \!\!\!\! Asynchronous but requires priority information for planning.}
\vspace{-2em}
\end{table}

The algorithm \textbf{decMPC}~\cite{toumieh_decentralized_2022} is decentralized, but it requires synchronicity and communication delays to be within a fixed planning period.
\textbf{decGroup}~\cite{hou_enhanced_2022} is a decentralized semi-asynchronous planner, which solves joint optimization as a group. 
decGroup cooperatively tackles the path-planning problem but implicitly assumes no communication delays. 
\textbf{ADPP}~\cite{cap_asynchronous_2013} is asynchronous\footnote{As in \cite{tordesillas_mader_2022}, we define asynchronous planning to be when the agent triggers trajectory planning independently without considering the planning status of other agents. 
However, ADPP~\cite{cap_asynchronous_2013} implements a prioritized asynchronous approach, meaning plannings are not fully independently triggered.} and decentralized, but it assumes perfect communication without delay. 
Our previous work \MADER{}~\cite{tordesillas_mader_2022} is asynchronous and decentralized but assumes no communication delays.
\textbf{EGO-Swarm}~\cite{zhou_ego-swarm_2020} also proposes a decentralized, asynchronous planner that requires agents to periodically broadcast a trajectory at a fixed frequency, and each agent immediately performs collision checks upon receiving the message. \EGOswarm{} is the first fully decentralized, asynchronous trajectory planner successfully demonstrating hardware experiments, yet it still suffers from a collision due to communication delays, as shown in Section~\ref{sec:sim}. 
\textbf{AsyncBVC}~\cite{senbaslar_asynchronous_2022} proposes an asynchronous decentralized trajectory planner that can guarantee safety even with communication delays.
However, the future trajectories are constrained by past separating planes, which can overconstrain the solution space and hence increase the conservatism. Also, they only presented simulation results with up to 4 agents, and no hardware experiments were implemented. In addition, it relies on discretization when solving the optimization problem, meaning that safety is only guaranteed on the discretization points. Our approach instead is able to guarantee safety in a continuous approach by leveraging the MINVO basis~\cite{tordesillas_minvo_2022}.

To address these shortcomings, we propose \textbf{Robust MADER} (\RMADER{}), a decentralized and asynchronous multiagent trajectory planner capable of generating collision-free trajectories in the presence of realistic communication delays.
As shown in Table~\ref{tab:state_of_the_art_comparison}, RMADER is the first approach to demonstrate decentralized, asynchronous trajectory planning robust to communication delays. RMADER builds on convex MADER, which is a modified version of the nonconvex MADER presented in our previous work~\cite{tordesillas_mader_2022} (more details are available in Appendix~\ref{sec:convex_nonconvex_MADER}).
Our contributions include:
\begin{enumerate}
  \item An algorithm that guarantees collision-free trajectory generation even with the existence of communication delays among vehicles. 
  \item Extensive simulations comparing our approach to state-of-the-art methods under communication delays that demonstrate \textbf{a 100\% success rate} of collision-free trajectory generation (see Table~\ref{tab:sim_compare}).
  \item Extensive set of decentralized hardware experiments using 6 UAVs, and achieving velocities up to $3.4$~m/s.
\end{enumerate}



\section{Trajectory Deconfliction} 

\begin{figure*}[t]
\centering
\begin{minipage}{.51\textwidth}
  \centering
  \resizebox{1.0\textwidth}{!}{%
    \begin{tikzpicture}
        [
        greenbox/.style={shape=rectangle, fill=opt_color, draw=black},
        bluebox/.style={shape=rectangle, fill=check_color, draw=black},
        redbox/.style={shape=rectangle, fill=recheck_color, draw=black},
        ]
        
        \newcommand\Ay{2.5}
        \newcommand\Axo{1}
        \newcommand\Axc{3}
        \newcommand\Axr{4.3}
        \newcommand\Axe{4.5}
        
        \newcommand\By{0.7}
        \newcommand\Bxo{2.3}
        \newcommand\Bxc{5.2}
        \newcommand\Bxr{6.5}
        
            \node[text=red] at (0.5,\Ay+0.2) {\scriptsize Agent A};
            \filldraw[fill=check_color, draw=black, opacity=0.2] (0,\Ay) rectangle (1.5,\Ay-0.3);
            \filldraw[fill=recheck_color, draw=black, opacity=0.2] (0.5,\Ay) rectangle (\Axo,\Ay-0.3);
            \filldraw[thick, fill=opt_color, draw=black] (\Axo,\Ay) rectangle (\Axc,\Ay-0.3);
            \filldraw[thick, fill=check_color, draw=black] (\Axc, \Ay) rectangle (\Axr, \Ay-0.3);
            \filldraw[thick, fill=recheck_color, draw=black] (\Axr, \Ay) rectangle (\Axe, \Ay-0.3);
            \filldraw[fill=opt_color, draw=black, opacity=0.2] (\Axe, \Ay) rectangle (\Axe+2.5, \Ay-0.3);
            \filldraw[fill=check_color, draw=black, opacity=0.2] (\Axe+2.5, \Ay) rectangle (\Axe+3.5, \Ay-0.3);
            \filldraw[fill=recheck_color, draw=black, opacity=0.2] (\Axe+3.5, \Ay) rectangle (\columnwidth, \Ay-0.3);
            \node[text=blue] at (0.5,\By+0.2) {\scriptsize Agent B};
            \filldraw[fill=check_color, draw=black, opacity=0.2] (0,\By) rectangle (\Bxo-0.5,\By-0.3);
            \filldraw[fill=recheck_color, draw=black, opacity=0.2] (\Bxo-0.5,\By) rectangle (\Bxo,\By-0.3);
            \filldraw[thick, fill=opt_color, draw=black] (\Bxo,\By) rectangle (\Bxc,\By-0.3);
            \filldraw[thick, fill=check_color, draw=black] (\Bxc, \By) rectangle (\Bxr, \By-0.3);
            \filldraw[thick, fill=recheck_color, draw=black] (\Bxr, \By) rectangle (\Bxr+0.2, \By-0.3);
            \filldraw[fill=opt_color, draw=black, opacity=0.2] (\Bxr+0.2, \By) rectangle (\Bxr+2.0, \By-0.3);
            \filldraw[fill=check_color, draw=black, opacity=0.2] (\Bxr+2.0, \By) rectangle (\columnwidth, \By-0.3);
        
        \draw[thick, densely dotted] (\Axe,-0.0) -- (\Axe,\Ay-0.3) node[] at (\Axe, -0.3) {\tiny $t_{\trajA{}{}}^A$};
            
        \draw[thick,->] (0,0) -- (\columnwidth,0) node[anchor=north east] {time};
        
        \draw[thick, ->, draw=red] (\Axe,\Ay-0.3) -- (\Axe,\Ay-1.2) node[midway,fill=white, text=red] {\tiny \trajA{}};
        \draw[thick, <-, draw=red] (\Bxc-0.3,\By) -- (\Bxc-0.3,\By+0.3)  node[anchor=south,text=black] {\tiny case 1};
        \draw[thick, <-, draw=red] (\Bxc+0.5,\By) -- (\Bxc+0.5,\By+0.3) node[anchor=south,text=black] {\tiny case 2};
        \draw[thick, <-, draw=red] (\Bxr+0.1,\By) -- (\Bxr+0.1,\By+0.3) node[anchor=south,text=black] {\tiny case 3};
        \draw[thick, <-, draw=red] (\Bxr+0.9,\By) -- (\Bxr+0.9,\By+0.3) node[anchor=south,text=black] {\tiny case 4};
        
        \node[font=\bfseries,right] at (\Axo,\Ay-0.15) {\tiny Optimization (O\textsubscript{A})};
        \node[font=\bfseries,right] at (\Axc,\Ay-0.15) {\tiny Check (C\textsubscript{A})};
        \node[font=\bfseries,right] at (\Axr-0.47,\Ay+0.15) {\tiny Recheck (R\textsubscript{A})};

        \node[font=\bfseries,right] at (\Bxo,\By-0.15) {\tiny O\textsubscript{B}};
        \node[font=\bfseries,right] at (\Bxc,\By-0.15) {\tiny C\textsubscript{B}};
        \node[font=\bfseries,right] at (\Bxr-0.1,\By-0.5) {\tiny R\textsubscript{B}};
        
        
        \node[color=gray] at (0.5,\Ay-0.15) {\scriptsize Prev. iter.};
        \node[color=gray] at (0.95\columnwidth,\Ay-0.15) {\scriptsize Next iter.};
        \node[color=gray] at (0.5,\By-0.15) {\scriptsize Prev. iter.};
        \node[color=gray] at (0.95\columnwidth,\By-0.15) {\scriptsize Next iter.};
    \end{tikzpicture}
    }
        \vspace*{-5mm}
    \captionof{figure}{MADER deconfliction: \AgentA{} solves \OStepA{} to find its optimal trajectory, constrained by other agents' trajectories. \AgentA{} then begins \CStepA{} to determine if that generated trajectory has any conflicts with trajectories received during \OStepA{}. Finally, in \RStepA{}, \AgentA{} checks if it received any trajectories during \CStep{}. The four cases shown in the figure correspond to different communication delays, resulting in \trajA{} being received by \AgentB{} at different times. Designed to guarantee safety when there is no communication delays, MADER also guarantees safety when \trajA{} arrives during \OStepB{} (Case 1) or \CStepB{} (Case 2), but could cause collisions if \trajA{} is received during/after (Case 3/4) \RStepB{}.
    }
  \label{fig:mader_deconfliction}
\end{minipage}%
\hspace{.01\textwidth}
\vspace{-1em}
\begin{minipage}{.45\textwidth}
  \centering
  \resizebox{1.0\textwidth}{!}{%
       \begin{tikzpicture}
       [
        greenbox/.style={shape=rectangle, fill=opt_color, draw=black},
        bluebox/.style={shape=rectangle, fill=check_color, draw=black},
         yellowbox/.style={shape=rectangle, fill=delaycheck_color, draw=black},
        ]
        
        \newcommand\Ay{2.5}
        \newcommand\Axo{1}
        \newcommand\Axc{3}
        \newcommand\Axr{4}
        \newcommand\Axe{5.5}
        
        \newcommand\By{0.7}
        \newcommand\Bxo{2.0}
        \newcommand\Bxc{4.7}
        \newcommand\Bxr{6.0}
        \newcommand\Bxe{7.5}
        
            \node[text=red] at (0.5,\Ay+0.2) {\scriptsize Agent A};
            \filldraw[fill=check_color, draw=black, opacity=0.2] (0,\Ay) rectangle (1.5,\Ay-0.3);
            \filldraw[fill=delaycheck_color, draw=black, opacity=0.2] (0.5,\Ay) rectangle (\Axo,\Ay-0.3);
            \filldraw[thick, fill=opt_color, draw=black] (\Axo,\Ay) rectangle (\Axc,\Ay-0.3);
            \filldraw[thick, fill=check_color, draw=black] (\Axc, \Ay) rectangle (\Axr, \Ay-0.3);
            \filldraw[thick, fill=delaycheck_color, draw=black] (\Axr, \Ay) rectangle (\Axe, \Ay-0.3);
            \filldraw[fill=opt_color, draw=black, opacity=0.2] (\Axe, \Ay) rectangle (\Axe+1.5, \Ay-0.3);
            \filldraw[fill=check_color, draw=black, opacity=0.2] (\Axe+1.5, \Ay) rectangle (\columnwidth, \Ay-0.3);
            \node[text=blue] at (0.5,\By+0.2) {\scriptsize Agent B};
            \filldraw[fill=check_color, draw=black, opacity=0.2] (0,\By) rectangle (\Bxo-0.5,\By-0.3);
            \filldraw[fill=delaycheck_color, draw=black, opacity=0.2] (\Bxo-0.5,\By) rectangle (\Bxo,\By-0.3);
            \filldraw[thick, fill=opt_color, draw=black] (\Bxo,\By) rectangle (\Bxc,\By-0.3);
            \filldraw[thick, fill=check_color, draw=black] (\Bxc, \By) rectangle (\Bxr, \By-0.3);
            \filldraw[thick, fill=delaycheck_color, draw=black] (\Bxr, \By) rectangle (\Bxe, \By-0.3);
            \filldraw[fill=opt_color, draw=black, opacity=0.2] (\Bxe, \By) rectangle (\columnwidth, \By-0.3);
        
        \draw[thick, densely dotted] (\Axr,0) -- (\Axr,\Ay-0.3) node[] at (\Axr, -0.25) {\tiny t\textsubscript{traj\textsubscript{A\textsubscript{new}}}};
        \draw[thick, densely dotted] (\Axe,0) -- (\Axe,\Ay-0.3) node[] at (\Axe, -0.25) {\tiny t\textsubscript{traj\textsubscript{A}}};
            
        \draw[thick,->] (0,0) -- (\columnwidth,0) node[anchor=north east] {time};
        
        \draw[thick, ->, draw=red] (\Axr,\Ay-0.3) -- (\Axr,\Ay-1.2) node[midway,fill=white, text=red] {\tiny \trajANew{}};
        \draw[thick, ->, draw=red] (\Axe,\Ay-0.3) -- (\Axe,\Ay-1.2) node[midway,fill=white, text=red] {\tiny \trajA{}};
        \draw[thick, <-, draw=red] (\Bxc-0.2,\By) -- (\Bxc-0.2,\By+0.3)  node[anchor=south,text=black] {\tiny case 1};
        \draw[thick, <-, draw=red] (\Bxc+0.35,\By) -- (\Bxc+0.35,\By+0.3) node[anchor=south,text=black] {\tiny case 2};
        \draw[thick, <-, draw=red] (\Bxr+0.4,\By) -- (\Bxr+0.4,\By+0.3) node[anchor=south,text=black] {\tiny case 3};
        \draw[thick, <-, draw=red] (\Bxe+0.4,\By) -- (\Bxe+0.4,\By+0.3) node[anchor=south,text=black] {\tiny case 4};
        
        \node[font=\bfseries,right] at (\Axo,\Ay-0.15) {\tiny O\textsubscript{A}};
        \node[font=\bfseries,right] at (\Axc,\Ay-0.15) {\tiny C\textsubscript{A}};
        \node[font=\bfseries,right] at (\Axr,\Ay-0.15) {\tiny DC\textsubscript{A}};

        \node[font=\bfseries,right] at (\Bxo,\By-0.15) {\tiny O\textsubscript{B}};
        \node[font=\bfseries,right] at (\Bxc,\By-0.15) {\tiny C\textsubscript{B}};
        \node[font=\bfseries,right] at (\Bxr,\By-0.15) {\tiny DC\textsubscript{B}};
        
        
        \node[color=gray] at (0.5,\Ay-0.15) {\scriptsize Prev. iter.};
        \node[color=gray] at (0.95\columnwidth,\Ay-0.15) {\scriptsize Next iter.};
        \node[color=gray] at (0.5,\By-0.15) {\scriptsize Prev. iter.};
        \node[color=gray] at (0.95\columnwidth,\By-0.45) {\scriptsize Next iter.};
        
    \end{tikzpicture}
    }
    
  \captionof{figure}{RMADER deconfliction: After \CStepA{} \AgentA{} keeps executing the trajectory from the previous iteration, traj\textsubscript{A\textsubscript{prev}}, while checking potential collisions of newly optimized trajectory, \trajANew{}. This is because \trajANew{} might have conflicts due to communication delays and need to be checked in \DCStepA{}, and traj\textsubscript{A\textsubscript{prev}} is ensured to be collision-free. If collisions are detected in either \CStepA{} or \DCStepA{}, \AgentA{} keeps executing \trajAPrev{} (i.e., $\text{\trajA{}}\leftarrow \text{\trajAPrev{}}$). If \DCStepA{} does not detect collisions, \AgentA{} broadcasts and starts implementing \trajANew{} (i.e., $\text{\trajA{}}\leftarrow \text{\trajANew{}}$).
    }
  \label{fig:rmader_deconfliction}
\end{minipage}
\end{figure*}

In MADER~\cite{tordesillas_mader_2022} and RMADER, UAVs plan trajectories asynchronously and broadcast the results to each other.  Each agent uses these trajectories as constraints in the optimization problem. Assuming no communication delays exist, safety can be guaranteed using our previous approach presented in MADER (summarized in Section~\ref{subsec:mader_deconfliction}). 
This safety guarantee, however, breaks when an agent's planned trajectory is received by other agents with some latency. Section~\ref{subsec:rmader_deconfliction} shows how RMADER guarantees safety even with communication delays. We use the definitions shown in Table~\ref{tab:delaydefinitions}.

\begin{table}[b]
    \vspace{-1.5em}
    \begin{centering}
    \caption{Definitions of the different delay quantities: Note that, by definition, $0\le\delayIntroduced\le\delayActual{}\le\delayActualMax{}$. See also Figs~\ref{fig:sim_actual_comm_delay} and~\ref{fig:comm_delay_on_centr} for the actual histogram of the delays in simulation and hardware, respectively.  }
    \renewcommand{\arraystretch}{2}
    \begin{centering}
    \begin{tabular}{>{\centering\arraybackslash}m{0.1\columnwidth} >{\arraybackslash}m{0.7\columnwidth}}
    \toprule 
    \delayActual{} & Actual communication delays among agents. \tabularnewline
    \hline 
    \delayActualMax{} & \makecell[l]{Possible maximum communication delay. } \tabularnewline
    \hline 
    \delayIntroduced{} & \makecell[l]{Introduced communication delay in simulations. } \tabularnewline
    \hline 
    \delayParameter{} & \makecell[l]{Length of \DelayCheckStep{} in RMADER. \\ To guarantee safety, $\delayActualMax{}\le\delayParameter{}$ must be satisfied.} \tabularnewline
    \bottomrule
    \end{tabular}
    \par\end{centering}
    \label{tab:delaydefinitions}
    \par\end{centering}
\end{table}



\subsection{MADER Deconfliction} \label{subsec:mader_deconfliction}
MADER~\cite{tordesillas_mader_2022} guarantees collision-free trajectories under ideal communication through the use of the planning stages shown in Fig.~\ref{fig:mader_deconfliction}.
An agent plans its initial trajectory during \OptimizationStep{} (\OStep{}), followed by \CheckStep{} (\CStep{}) to ensure its plan does not lead to a collision.
Finally, \RecheckStep{} (\RStep{}) is used to check if the agent received any trajectory updates from other agents during \CStep{} - if so, an agent starts over planning at \OStep{}.
Although MADER does not have explicit safety guarantees in the presence of communication delays, its trajectories are still collision free for cases 1 and 2 shown in Fig.~\ref{fig:mader_deconfliction}. However,  collisions may occur in cases 3 and 4 of Fig.~\ref{fig:mader_deconfliction}.
These four cases are summarized in Fig.~\ref{fig:mader_deconfliction} and Table~\ref{tab:safety_guarantees_delays}.

\subsection{Robust MADER Deconfliction} \label{subsec:rmader_deconfliction}

To achieve robustness to communication delays, we replace the \RecheckStep{} with \DelayCheckStep{} (\DCStep{}), where each agent repeatedly checks if its newly optimized trajectory conflicts with other agents' trajectories.
If an agent detects conflicts during \DCStep{}, it discards the new trajectory and starts another \OStep{} while executing its previous trajectory. 
If no collisions are detected in \DCStep{}, it starts executing the new trajectory. 
To guarantee collision-free trajectory generation, \DCStep{} needs to be longer than the possible longest communication delay (i.e., \NeccessaryCond{}). 
That way, an agent can always keep at least one collision-free trajectory. 
It could, however, not be ideal for introducing such a long \delayParameter{}, and therefore, in Section~\ref{sec:sim} we also tried $\delta_{\text{DC}}<\delta_{\text{max}}$ and measure its performance. 
Fig.~\ref{fig:rmader_deconfliction} shows how RMADER deals with communication delays.
Furthermore, Table~\ref{tab:safety_guarantees_delays} shows all the possible cases in which communication delays could occur and how these are handled by RMADER to generate collision-free trajectories even with communication delays.
The pseudocode of RMADER deconfliction is given in Algorithm~\ref{alg:rmader}. First, Agent~B runs \OStep{} to obtain \trajBNew{} and broadcasts it if \CStep{} is satisfied 
(Line~\ref{line:broadcast_traj_B_new}). This \CStep{} aims to determine if \trajBNew{} has any conflicts with trajectories received in \OStep{}. Then, Agent~B commits either \trajBPrev{} or \trajBNew{} - if \DCStep{} detects conflicts, Agent~B commits to \trajBPrev{} (Line~\ref{line:DC_not_satisfied}), and if \DCStep{} detects no conflicts, \trajBNew{} (Line~\ref{line:DC_satisfied}). This committed trajectory is then broadcast to the other agents (Line~\ref{line:broadcast_traj_B}). Table~\ref{tab:mader_vs_rmader} highlights the differences between MADER and RMADER.

\begin{table}
\caption{Safety guarantees under communication delays: Depending on when \trajA{} is received by Agent B, the deconfliction takes place at different stages. MADER does not guarantee safety if \trajA{} is received during \RStepB{} or during the following iteration, while RMADER guarantees safety in all the cases. Note that in RMADER, if Agent~B does not receive \trajA{} by the end of \DCStepB{}, then the deconfliction is performed by Agent A (specifically, in \CStepA{} or \DCStepA{}) and not by Agent B. Agent A will use \trajBNew{} and/or \trajB{} for this.}
\label{tab:safety_guarantees_delays}
\begin{centering}
\renewcommand{\arraystretch}{1.2}
\resizebox{1.0\columnwidth}{!}{
\setlength\extrarowheight{0.3em}
\begin{tabular}{>{\centering}m{0.3\columnwidth} || >{\centering}m{0.2\columnwidth} >{\centering}m{0.2\columnwidth} >{\centering}m{0.2\columnwidth} >{\centering}m{0.2\columnwidth}}
\toprule
\multicolumn{1}{>{\centering}m{0.1\columnwidth}||}{} & \multicolumn{4}{c}{\textbf{MADER}} \tabularnewline[0.2em]
\hline
\textbf{When \trajA{} received}  & \makecell{\OStepB{} \\ (Case 1)} & \makecell{\CStepB{} \\ (Case 2)} & \makecell{\RStepB{} \\ (Case 3)} & \makecell{\textbf{Next iter.} \\ (Case 4)} \tabularnewline[0.4em]
\hline 
\textbf{\trajA{} deconflicted? When?} & \makecell{\YesGreen{} \\ \CStepB{}}  & \makecell{\YesGreen{} \\ \RStepB{}}  & \NoRed{}  & \NoRed{} \tabularnewline
\bottomrule
\end{tabular}
}
\vspace{0.2cm}

\resizebox{1.0\columnwidth}{!}{
\setlength\extrarowheight{0.3em}
\begin{tabular}{>{\centering}m{0.3\columnwidth} || >{\centering}m{0.2\columnwidth} >{\centering}m{0.2\columnwidth} >{\centering}m{0.2\columnwidth} >{\centering}m{0.2\columnwidth}}
\toprule
\multicolumn{1}{>{\centering}m{0.1\columnwidth}||}{} & \multicolumn{4}{c}{\textbf{RMADER}}\tabularnewline[0.2em]
\hline
\textbf{When \trajANew{}/\trajA{} received}  & \makecell{\OStepB{} \\ (Case 1)} & \makecell{\CStepB{} \\ (Case 2)} & \makecell{\DCStepB{} \\ (Case 3)} & \makecell{\textbf{Next iter.} \\ (Case 4)} \tabularnewline
\hline 
\textbf{\trajANew{}/\trajA{} deconflicted? When?} & \makecell{\YesGreen{} \\ \CStepB{}} & \makecell{\YesGreen{} \\ \DCStepB{}}  & \makecell{\YesGreen{} \\ \DCStepB{}}  & \makecell{\YesGreen{} \\ \CStepA{} or \DCStepA{}} \tabularnewline
\bottomrule
\end{tabular}
}

\par\end{centering}
\end{table}

\begin{algorithm}
    \begin{algorithmic}[1]
    \Require traj\textsubscript{B}, a feasible trajectory
    \While{not goal reached}
        \State \trajBNew{} $=$ \textproc{\OptimizationStep{}()} \label{line:traj_opt}
        \If{\textproc{\CheckStep}(\trajBNew{}) $==$ False} 
            \State Go to Line 2
        \EndIf
        \State Broadcast \trajBNew{} \label{line:broadcast_traj_B_new}
        \If{\textproc{\DelayCheckStep{}}(\trajBNew{}) $==$ False} 
            \State \trajB{} $\leftarrow$ \trajBPrev{}, and go to Line~\ref{line:broadcast_traj_B}\label{line:DC_not_satisfied}
        \EndIf
        \State \trajB{} $\leftarrow$ \trajBNew{} \label{line:DC_satisfied}
        \State Broadcast \trajB{} \label{line:broadcast_traj_B}
    \EndWhile
    \end{algorithmic}
    \caption{Robust MADER - Agent B}
    \label{alg:rmader}
\end{algorithm}




\begin{table}
\begin{centering}
\caption{\centering Differences between MADER and RMADER}
\label{tab:mader_vs_rmader}
\renewcommand{\arraystretch}{1.1}
\resizebox{1.0\columnwidth}{!}{
\scriptsize
\begin{tabular}{>{\centering\arraybackslash}m{0.4\columnwidth} >{\centering\arraybackslash}m{0.4\columnwidth}}
\toprule
\textbf{MADER} & \textbf{RMADER}\tabularnewline
\hline 
\hline 
Upon successful \CStep{} and \RStep{}, the newly optimized trajectory is broadcast to other agents 
& Upon successful \CStep{}, \trajJNew{} is broadcast. After \DCStep{}, the committed trajectory \trajJ{} (which is either \trajJNew{} and \trajJPrev{} depending on whether \DCStep{} is satisfied or not) is broadcast \tabularnewline
\hline 
\RStep{} is a Boolean check to see if the agent received traj.
in \CStep{} & \DCStep{} is a sequence of collision checks\tabularnewline
\hline 
\RStep{} is very short & \DCStep{} lasts \delayParameter{} seconds\tabularnewline
\bottomrule
\end{tabular}
}
\par\end{centering}
\vspace{-2em}
\end{table}

\begin{algorithm}
    \newcommand{\algorithmicbreak}{\textbf{break}}
    \begin{algorithmic}[1] 
    \Function {\textproc{\DelayCheckStep{}}}{\trajBNew{}}
        \For {\delayParameter{} seconds}
            \If{\trajBNew{} collides with any trajectory in $\mathcal{Q}_B$}    
                \State \Return False
             \EndIf
         \EndFor
    \State \Return True
    \EndFunction
    \end{algorithmic}
    \caption{Delay Check - Agent B}\label{alg:pess_delaycheck}
\end{algorithm}


\newcommand{\QB}{\ensuremath{\mathcal{Q}_B}}

\begin{figure}
    \centering
    \includegraphics[width=0.6\columnwidth]{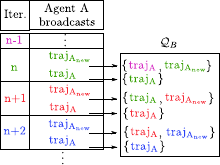}
    \setlength{\belowcaptionskip}{-0.5em}
    \caption{Agent~B stores in \QB{} the last committed trajectory of Agent~A. It will also contain the newly optimized trajectory \trajANew{} while the new committed trajectory has still not been received by Agent~B. }
    \label{fig:QB_definition}
\end{figure}

Agent~B stores the trajectories received from other agents in a set \QB{}; for example, for each Agent~J, Agent~B will store the committed trajectory of Agent~J, \trajJ{}, and possibly the newly optimized trajectory \trajJNew{} if any new committed trajectory has still not been broadcast. Figure~\ref{fig:QB_definition} shows the way Agent B stores trajectories from Agent A. 
\QB{} is used in \OStep{}, \CheckStep{}, and \DCStep{},  where Agent~B checks for collision against all the trajectories stored in \QB{}. Note that \QB{} is updated in parallel with \OStep{} and \DCStep{}. 
At the beginning of \OStepB{}, an agent generates an optimal trajectory, \trajBNew{}, using all the trajectories stored in \QB{} as constraints, and then, Agent B checks for \trajBNew{}'s potential collisions against \QB{}, which is updated in optimization. 
Finally, Agent~B repeatedly checks for collisions against \QB{} in \DCStepB{}, which lasts for \delayParameter{}. 
Note that \CStep{} is not necessary to guarantee safety in RMADER - \OStep{} followed by \DCStep{} alone can generate collision-free trajectories as long as \NeccessaryCond{} holds. Though \CStep{} detects collisions before broadcasting any (possibly conflicted) trajectories and allows an agent to start another \OStep{}, which prevents unnecessary communication.

\section{Simulation Results}\label{sec:sim}


\colorlet{color0ms}{blue}
\colorlet{color50ms}{OliveGreen}
\colorlet{color100ms}{RawSienna}
\colorlet{color200ms}{Magenta}
\colorlet{color300ms}{Orange}

\newcommand{\threec}[3]{\ensuremath{\textcolor{color0ms}{#1}|\textcolor{color50ms}{#2}|\textcolor{color100ms}{#3}}}
\newcommand{\fivec}[5]{\ensuremath{\textcolor{color0ms}{#1}|\textcolor{color50ms}{#2}|\textcolor{color100ms}{#3}|\textcolor{color200ms}{#3}|\textcolor{color300ms}{#5}}}
\newcommand{\threecb}[3]{\ensuremath{\textcolor{color0ms}{\textbf{#1}}|\textcolor{color50ms}{\textbf{#2}}|\textcolor{color100ms}{\textbf{#3}}}}
\newcommand{\fivecb}[5]{\ensuremath{\textcolor{color0ms}{\textbf{#1}}|\textcolor{color50ms}{\textbf{#2}}|\textcolor{color100ms}{\textbf{#3}}|\textcolor{color200ms}{\textbf{#3}}|\textcolor{color300ms}{\textbf{#3}}}}

\begin{table*}[!h]
    \renewcommand{\arraystretch}{2}
    \caption{\centering Cases \textcolor{color0ms}{$\delayIntroduced{}=0$~ms}, \textcolor{color50ms}{$\delayIntroduced{}=50$~ms}, \textcolor{color100ms}{$\delayIntroduced{}=100$~ms}, (see  Fig.~\ref{fig:sim_actual_comm_delay} for actual message delays). The bold values represent the case where \NeccessaryCond{}, which is the necessary condition to ensure safety.}
    \label{tab:sim_compare}    \renewcommand{\arraystretch}{1.6}
    \centering
    \resizebox{\textwidth}{!}{
    \begin{tabular}{  >{\centering\arraybackslash}m{0.1\textwidth} | >{\centering\arraybackslash}m{0.2\textwidth} | >{\centering\arraybackslash}m{0.1\textwidth} | >{\centering\arraybackslash}m{0.12\textwidth} |  >{\centering\arraybackslash}m{0.13\textwidth} | >{\centering\arraybackslash}m{0.18\textwidth} | >{\centering\arraybackslash}m{0.12\textwidth} | >{\centering\arraybackslash}m{0.12\textwidth}  }
        \toprule
        \multirow{2}{0.1\textwidth}{\centering Method} & \multirow{2}{0.08\textwidth}{\centering \delayParameter{} [ms]} & \multirow{2}{0.06\textwidth}{\centering Collision [\%]} & \arraybackslash  \arraybackslash \multirow{2}{0.1\textwidth}{\centering Avg number \\ of stops} & \multirow{2}{0.06\textwidth}{\centering \intaccelsquared{} [\SI{}{\m^2/\s^3}]} & \multirow{2}{0.06\textwidth}{\centering \intjerksquared{} [\SI{}{\m^2/\s^5}]} & \multicolumn{2}{c}{\centering Travel Time [s]}  \tabularnewline
         & & & & & & Avg & Max \tabularnewline
        \hline \hline
        \SlowEGOswarm{} & N/A & \threec{14}{25}{22} & \threec{0}{0}{0} & \threec{109.8}{113.2}{113.5} & \threec{15388.2}{15491.5}{15486.2} & \threec{11.65}{11.67}{11.76} & \threec{11.93}{11.99}{12.97}  \tabularnewline
        \hline \hline
        \EGOswarm{} & N/A & \threec{64}{84}{84} & \threec{0.004}{0.0}{0.01} & \threec{662.5}{700.7}{787.9} & \threec{90721.9}{94611.9}{104160.3} & \threec{7.19}{7.24}{7.28} & \threec{7.38}{7.51}{7.63} \tabularnewline
        \hline \hline
        \MADER{} \textbf{(convex)} & N/A & \threec{15}{38}{42} & \threec{0.0}{0.001}{0.0} & \threec{78.09}{74.19}{74.74} & \threec{1595.9}{1643.6}{1638.5} & \threec{6.28}{6.25}{6.26} & \threec{7.15}{7.35}{7.04} \tabularnewline
        \hline \hline
        \arraybackslash \multirow{2}{=}[-1.2em]{\centering \RMADER{} \textbf{(proposed)}} & \makecell{\threecb{100}{130}{200} \\ ($>$100th percentile of \delayActual{} \\ and \NeccessaryCond{} holds)} & \multirow{1}{=}[-0.5em]{\centering \threecb{0}{0}{0}} & \multirow{1}{=}[-0.5em]{\centering \threecb{0.46}{0.347}{1.751}} & \multirow{1}{=}[-0.5em]{\centering \threecb{127.7}{147.9}{190.5}} & \multirow{1}{=}[-0.5em]{\centering \threecb{2939.4}{3712.4}{5942.1}} & \multirow{1}{=}[-0.5em]{\centering \threecb{7.28}{7.95}{10.35}} & \multirow{1}{=}[-0.5em]{\centering \threecb{8.41}{8.80}{11.91}} \tabularnewline
        \cline{2-8}
        & \makecell{\threec{25}{56}{105} \\  ($\approx$75th percentile of \delayActual{} \\ so \NeccessaryCond{} does not hold)} & \multirow{1}{=}[-0.5em]{\centering \threec{0}{0}{0}} & \multirow{1}{=}[-0.5em]{\centering \threec{0.001}{0.007}{0.086}} & \multirow{1}{=}[-0.5em]{\centering \threec{99.52}{112.0}{137.7}} & \multirow{1}{=}[-0.5em]{\centering \threec{1844.3}{2142.3}{3056.2}} & \multirow{1}{=}[-0.5em]{\centering \threec{6.80}{6.87}{7.30}} & \multirow{1}{=}[-0.5em]{\centering \threec{7.66}{8.02}{8.89}} \tabularnewline
        \bottomrule 
    \end{tabular}
    }
    \vspace{-1em}
\end{table*}

\begin{figure}
    \centering
    \subfloat[For Agent~J, the colored trajectory is the committed (safety-guaranteed) trajectory (\trajJ{}), and the grey trajectory is the newly optimized trajectory (\trajJNew{}).\label{fig:circle_traj_new}]{\includegraphics[width=0.49\columnwidth]{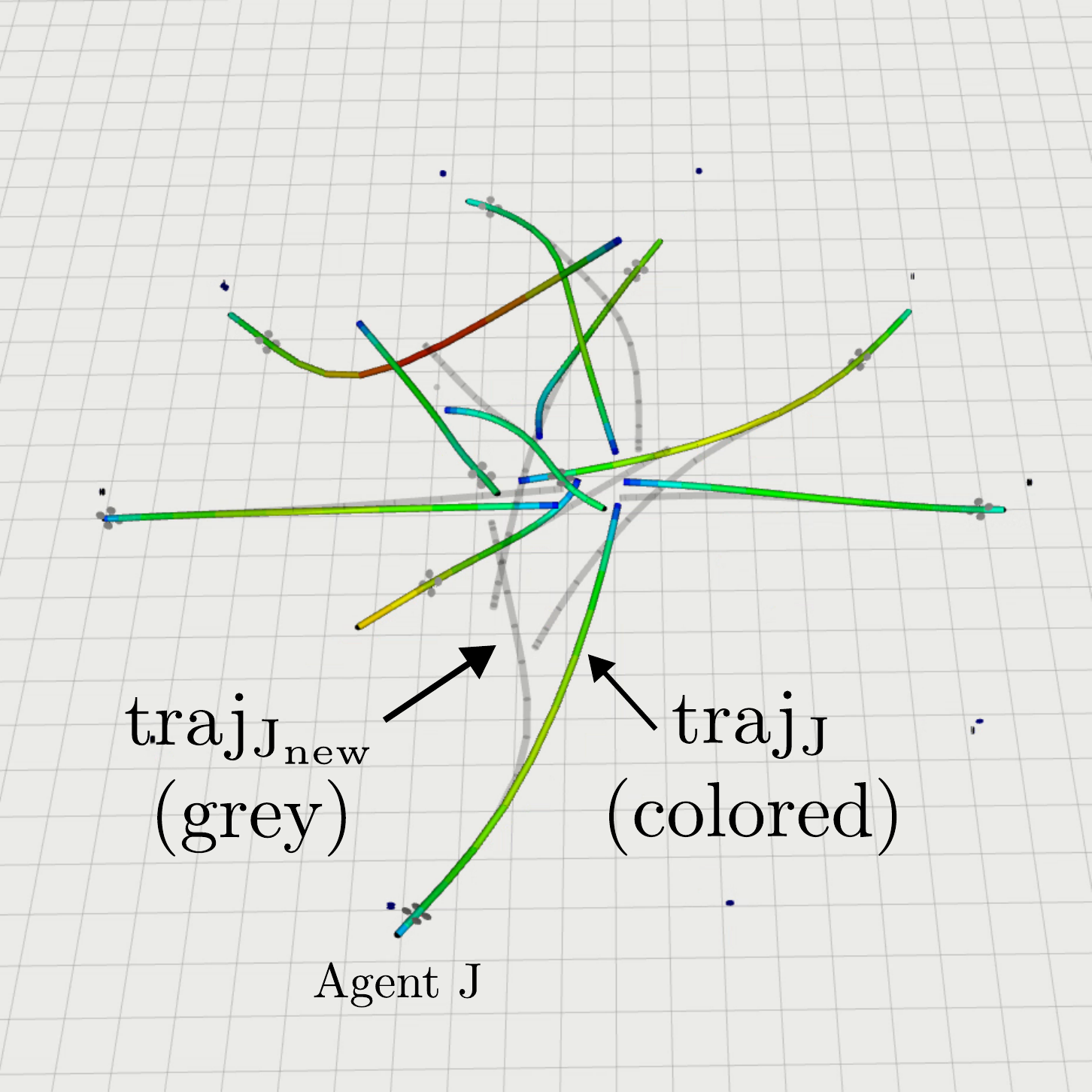}}
    \hfill
    \subfloat[Actual trajectories flown by the agents. All 10 agents successfully swap their positions in a circle configuration. \label{fig:circle_traj_history}]{\includegraphics[width=0.49\columnwidth]{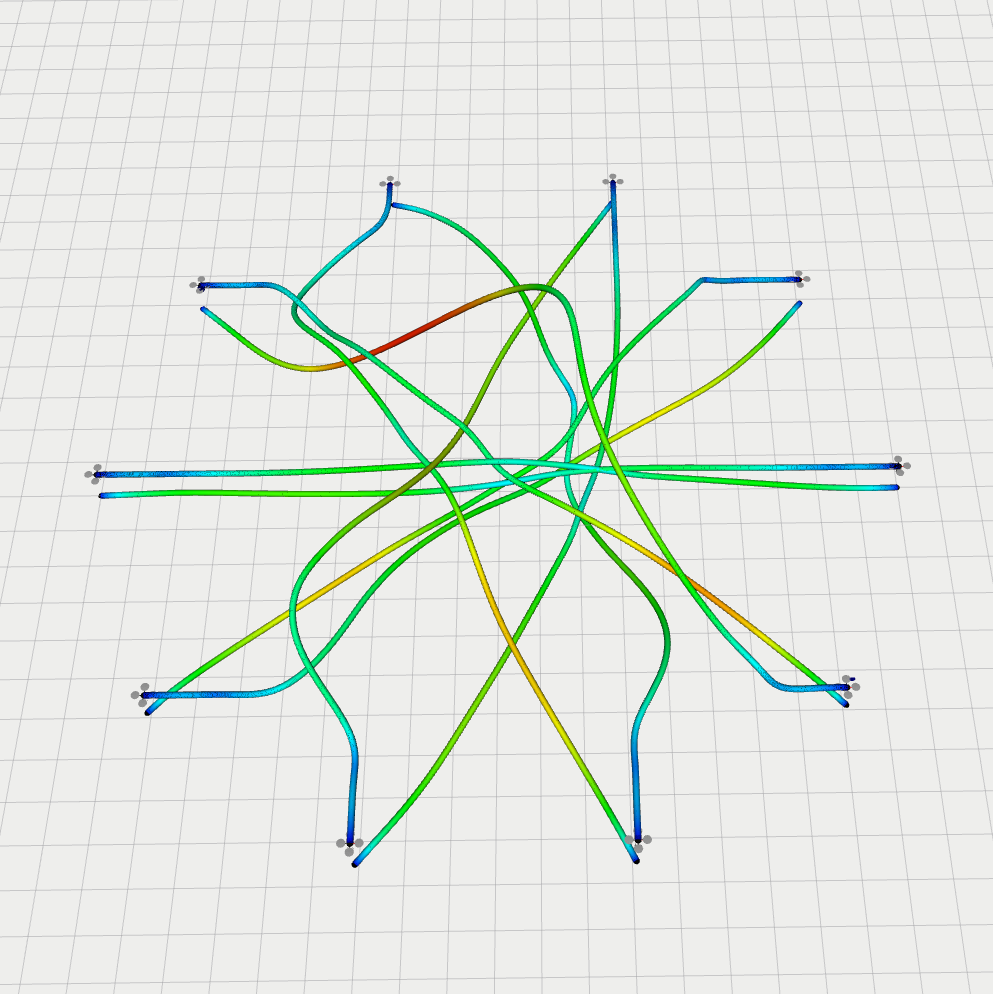}}
    \setlength{\belowcaptionskip}{-2em}
    \caption{10 agents employing RMADER exchange their positions in a circle of radius $20$~m. In the colored trajectories, red represents a high speed while blue denotes a low speed.
    }
    \label{fig:rmader_sim1}
\end{figure}

\begin{figure*}
    \centering
    \subfloat[Collision-free Trajectory Rate \label{fig:sim_collision_free_traj_rate}]{\includegraphics[width=0.33\textwidth]{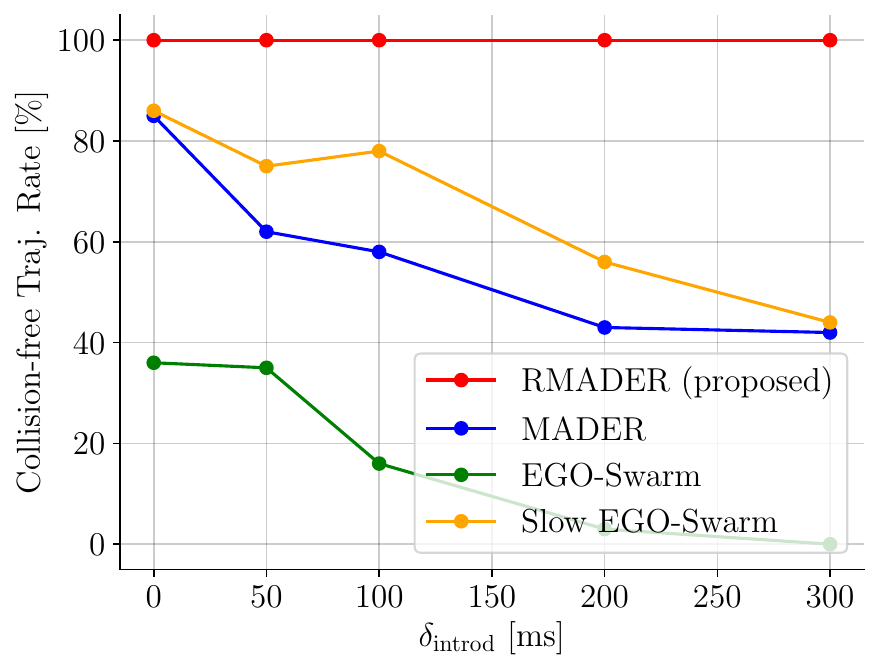}}
    \hfill
    \subfloat[Travel Time - shaded parts indicate its maximum and minimum value. \label{fig:sim_completion_time}]{\includegraphics[width=0.33\textwidth]{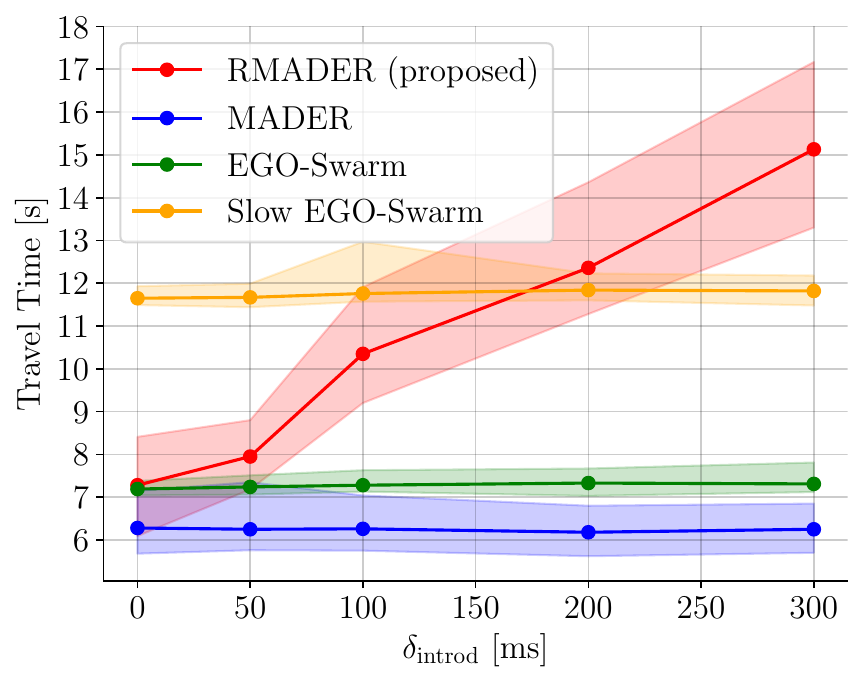}}
    \hfill
    \subfloat[Number of Stops \label{fig:sim_traj_smooth_acc}]{\includegraphics[width=0.33\textwidth]{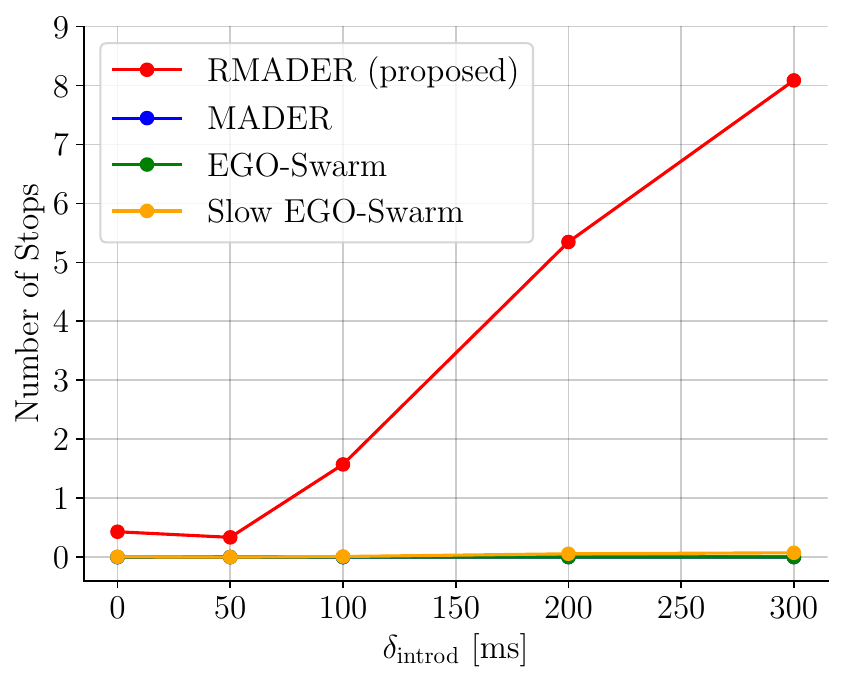}}
    
    \subfloat[Trajectory Smoothness (Acceleration) \label{fig:sim_traj_smooth_acc}]{\includegraphics[width=0.33\textwidth]{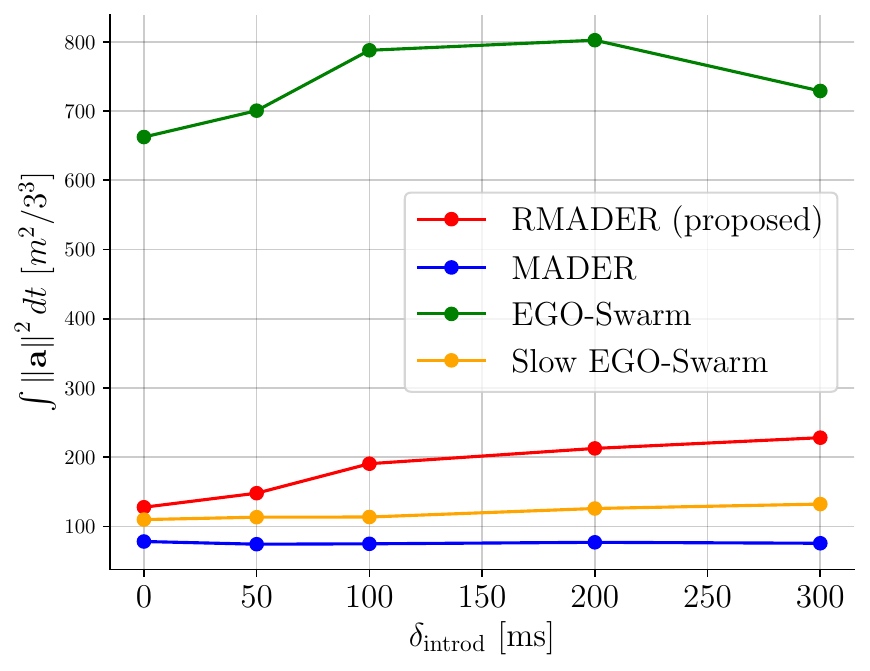}}
    \hfill
    \subfloat[Trajectory Smoothness (Jerk) \label{fig:sim_traj_smooth_jer}]{\includegraphics[width=0.33\textwidth]{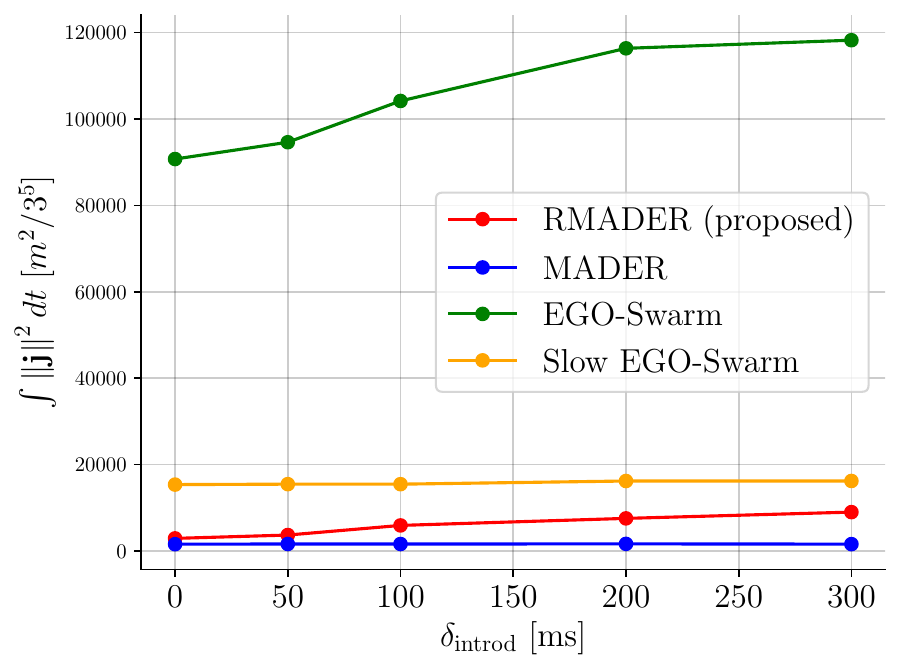}}
    \hfill
    \subfloat[Distribution of \delayActual{} in simulations. Due to computer’s computational limits, messages do not travel instantly.\label{fig:sim_actual_comm_delay}]{\includegraphics[width=0.33\textwidth]{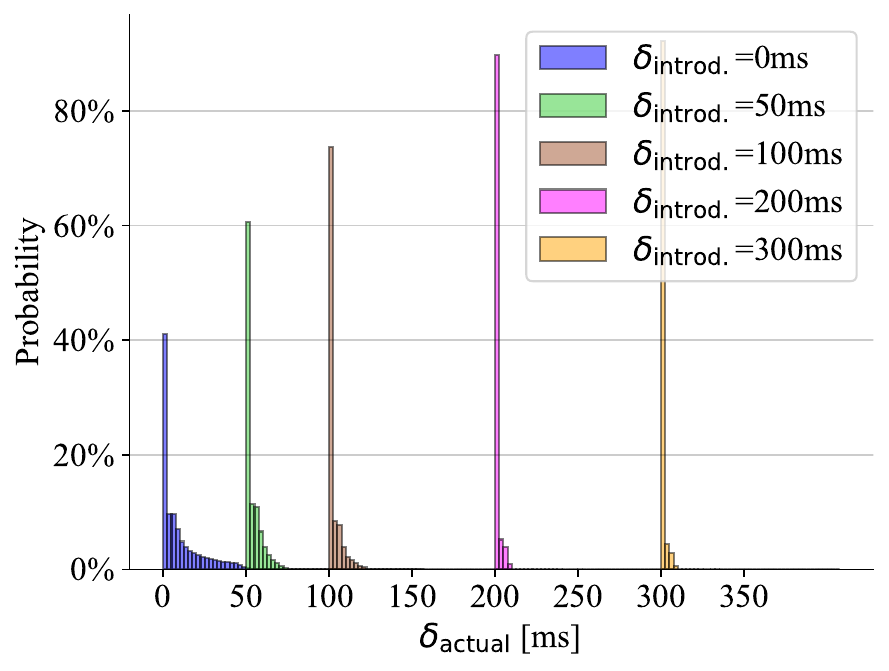}}
    \caption{100 Flight Simulation Results: Fig.~\ref{fig:sim_collision_free_traj_rate} shows RMADER generates collision-free trajectory at 100\%, while other state-of-the-art approaches fail when communication delays are introduced. 
    To maintain collision-free trajectory generation, RMADER periodically occupies two trajectories, and other agents need to consider two trajectories as a constraint, which could lead to conservative plans - longer \emph{Travel Time} and more \emph{Avg. Number of Stops}. 
    This is a trade-off between safety and performance. MADER reports a few collided trajectory because \delayActual{} $>$ \SI{0}{\ms}.}
    \label{fig:sim_simulation_summary}
    \vspace{-1em}
\end{figure*}


We tested Slow EGO-Swarm, EGO-Swarm~\cite{zhou_ego-swarm_2020}, MADER~\cite{tordesillas_mader_2022}, and RMADER (proposed) on a \texttt{general-} \texttt{purpose-N2} Google Cloud instance with 32 Intel Core i7s. In each scenario, we conducted \textbf{100 simulations} with 10 agents positioned in a \SI{10}{\m} radius circle, exchanging positions diagonally as shown in Fig.~\ref{fig:rmader_sim1}.
Note that this paper convexified MADER optimization problem as detailed in Appendix~\ref{sec:convex_nonconvex_MADER}, and we used the convex optimization problem for both MADER and RMADER.
The maximum dynamic limits (velocity, acceleration, and jerk) for these algorithms are set to \SI{10}{\m/\s}, \SI{20}{\m/\s^2}, and \SI{30}{\m/\s^3}. EGO-Swarm carries out a sequential startup - agents commits their first trajectory in a pre-determined order to avoid unnecessary trajectory conflicts. We also introduced \SI{0.25}{\second}-apart startup into MADER and RMADER. 

Slow EGO-Swarm is EGO-Swarm with smaller dynamic limits. 
We first tested EGO-Swarm with default parameters provided in \cite{zhou_ego-swarm_2020} and saw a significant number of conflicts. 
Therefore we increased the weights of the collision costs in EGO-Swarm's cost function up to 1000 (we tried more than 1000, but it did not change the results) while other weights (s.t. trajectory feasibility) are on the order of single digits; however, we still observed collisions (as seen in the second row of Table~\ref{tab:sim_compare}). 
We thus decreased the maximum velocity and acceleration of EGO-Swarm down to \SI{5}{\m/\s} and \SI{10}{\m/\s^2}, which we define as Slow EGO-Swarm.

Although we introduced a fixed $\delayIntroduced$ for simulated communications, \delayActual{} can be larger due to the simulation computer's computational limitations. 
The communication delays observed in simulation are shown in Fig.~\ref{fig:sim_actual_comm_delay} for five nominal values of $\delayIntroduced$ ($\delayIntroduced$ = 0, 50, 100, 200, and 300 \SI{}{\ms}).
As long as \NeccessaryCond{} holds, RMADER can generate collision-free trajectories. 

Table~\ref{tab:sim_compare} and Fig.~\ref{fig:sim_simulation_summary} showcase each approach's performance in simulations. RMADER was implemented in the case of (1) \NeccessaryCond{} (\delayParameter{} $>$ 100th percentile of \delayActual{}) and (2) \NotNeccessaryCond{} (\delayParameter{}~$\approx$~75th percentile of \delayActual{}). When \NeccessaryCond{} holds, collision-free trajectory planning is guaranteed, and therefore RMADER generates 0 collisions for all the \delayIntroduced{}, while other approaches suffer collisions. As expected, the longer \delayIntroduced{} more collisions Slow EGO-Swarm, EGO-Swarm, and MADER generate. In the case of (2) \NotNeccessaryCond{}, although safety is not theoretically guaranteed, since \delayParameter{} is long enough, RMADER succeeds to generate collision-free trajectories. Note that Case (2) could have collisions in case agents have conflicted trajectories and their trajectories fall into the rest of $\approx$25\%.

It is also worth mentioning that RMADER's robustness to communication delays is obtained by layers of conflict checks and agents periodically occupying two trajectory spaces, which can result in generating conservative trajectories and trading off UAV performance. 
\emph{Avg. Number of Stops} in Table~\ref{tab:sim_compare}, for instance, suggests more stoppage than other approaches. As \intaccelsquared{} and \intjerksquared{} show RMADER's trajectories are less smooth than MADER, and RMADER takes longer \emph{Travel Time} than others (Slow EGO-Swarm takes more but that is because of its smaller dynamic limits, and therefore a direct comparison is not fair).  



\section{Hardware Experiments}





\begin{table}
\caption{RMADER hardware experiments: Maximum velocity and flight distance (sum of the distances of each UAV) in
the five experiments}
\label{tab:max_vel_flight_dist_rmader}
\begin{centering}
\renewcommand{\arraystretch}{1.2}
\resizebox{1.0\columnwidth}{!}{
\begin{tabular}{ c c c c c c }
\toprule
 & \textbf{Exp. 1} & \textbf{Exp. 2} & \textbf{Exp. 3} & \textbf{Exp. 4} & \textbf{Exp.5}\tabularnewline
\hline 
\hline 
\textbf{Max vel. [m/s]} & 2.6 & 2.7 & 3.4 & 2.8 & 2.7\tabularnewline
\hline 
\textbf{Flight distance [m]} & 268.7 & 330.7 & 309.3 & 354.0 & 351.6\tabularnewline
\bottomrule
\end{tabular}}
\par\end{centering}
\vspace{-0.5cm}
\end{table}

\begin{figure}
    \centering
    \includegraphics[width=0.7\columnwidth]{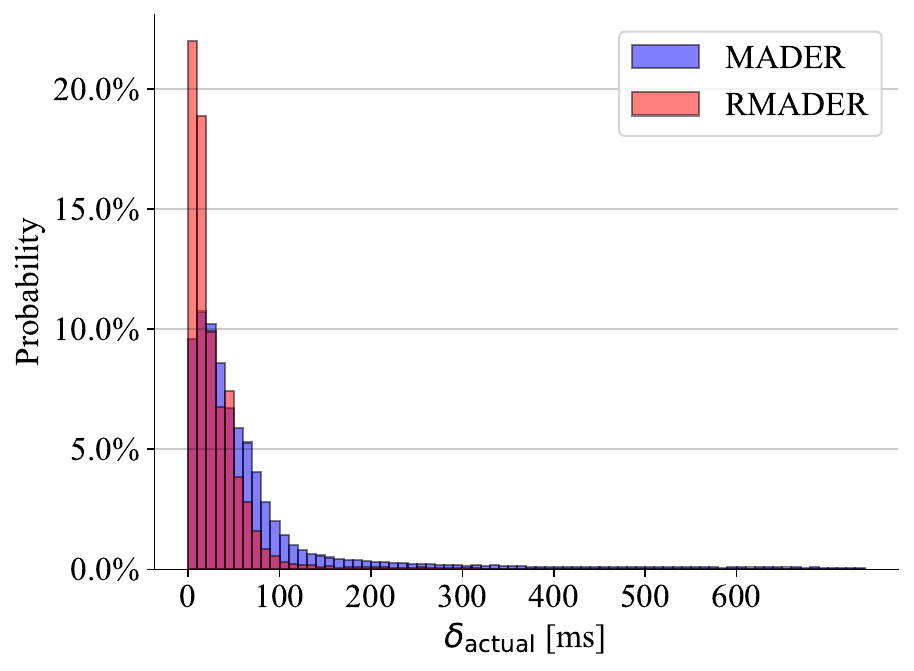} 
    \setlength{\belowcaptionskip}{-0.5em}
    \caption{Distribution of \delayActual{} in hardware experiments: Both MADER and RMADER were tested in 5 flight experiments. Compared to simulations (see the case $\delayIntroduced{} = $ \SI{0}{ms} in Fig. \ref{fig:sim_actual_comm_delay}), \delayActual{} is much larger in hardware.}
    \label{fig:comm_delay_on_centr}
\end{figure}

\begin{figure}[!htbp]
    \subfloat{\includegraphics[width=0.49\columnwidth, height=0.14\textheight]{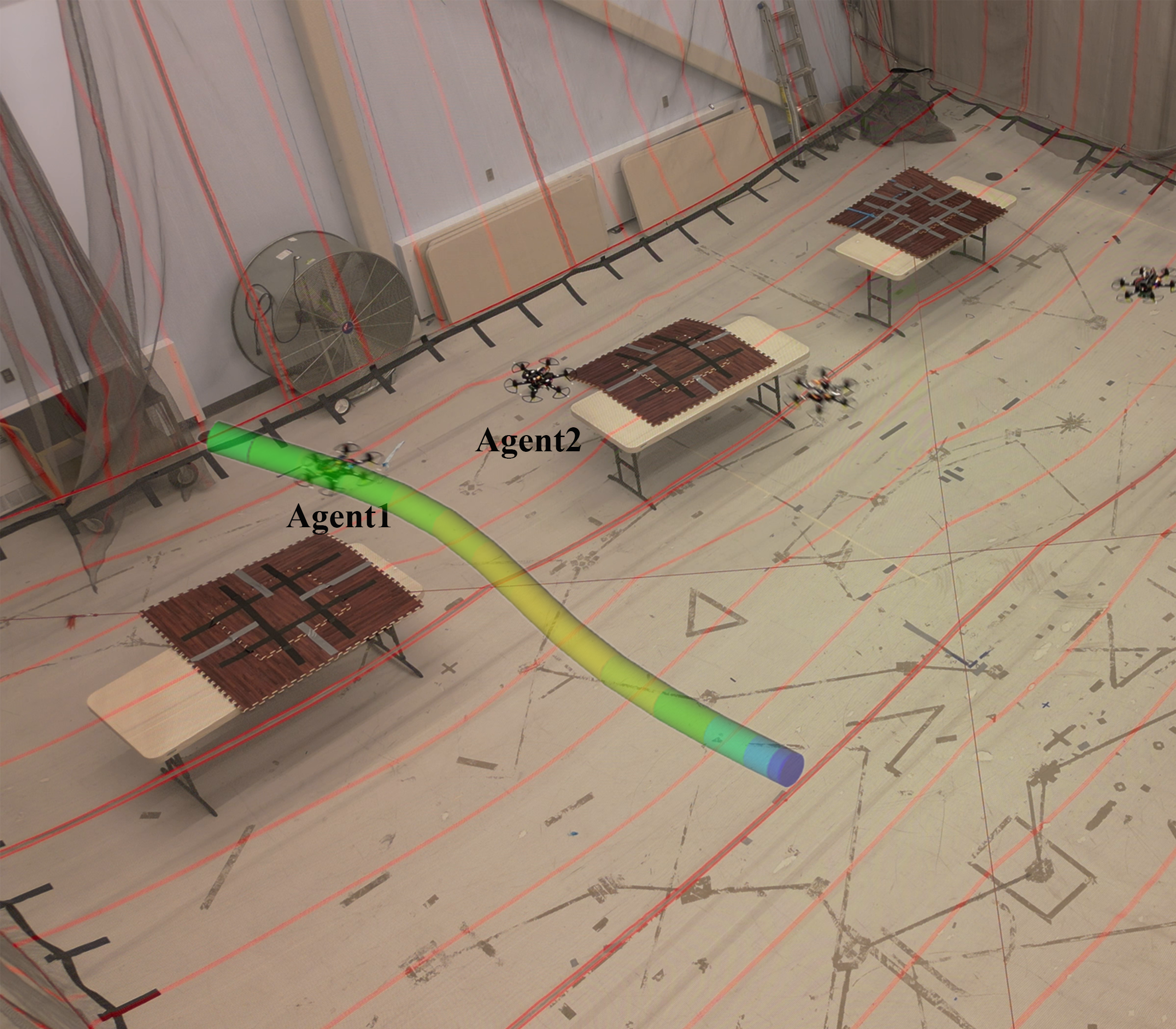}}
    \subfloat{\includegraphics[width=0.49\columnwidth, height=0.14\textheight]{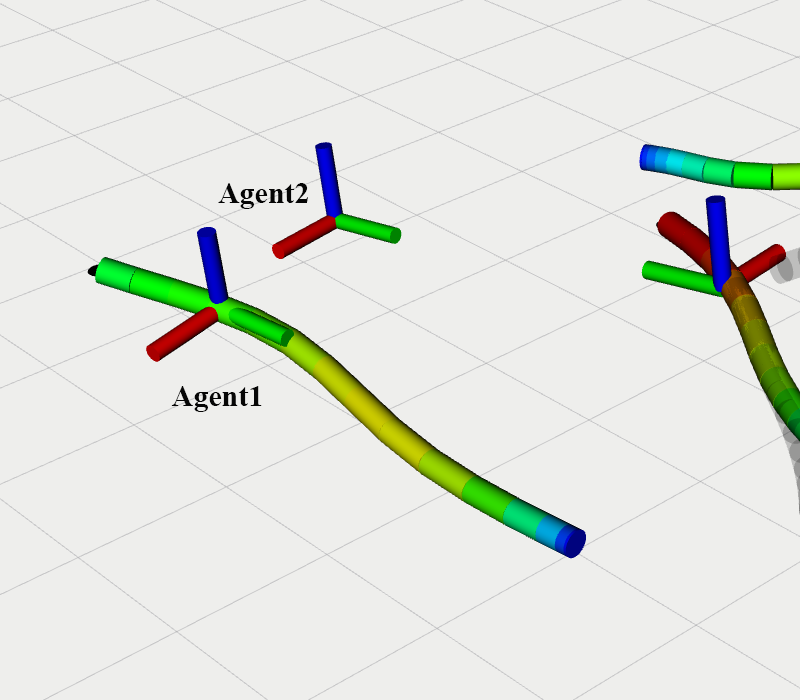}}
    \\ {\scriptsize $t=$ \SI{0}{\s}: Agent 1 is following its trajectory} \\[1em]
    \subfloat{\includegraphics[width=0.49\columnwidth, height=0.14\textheight]{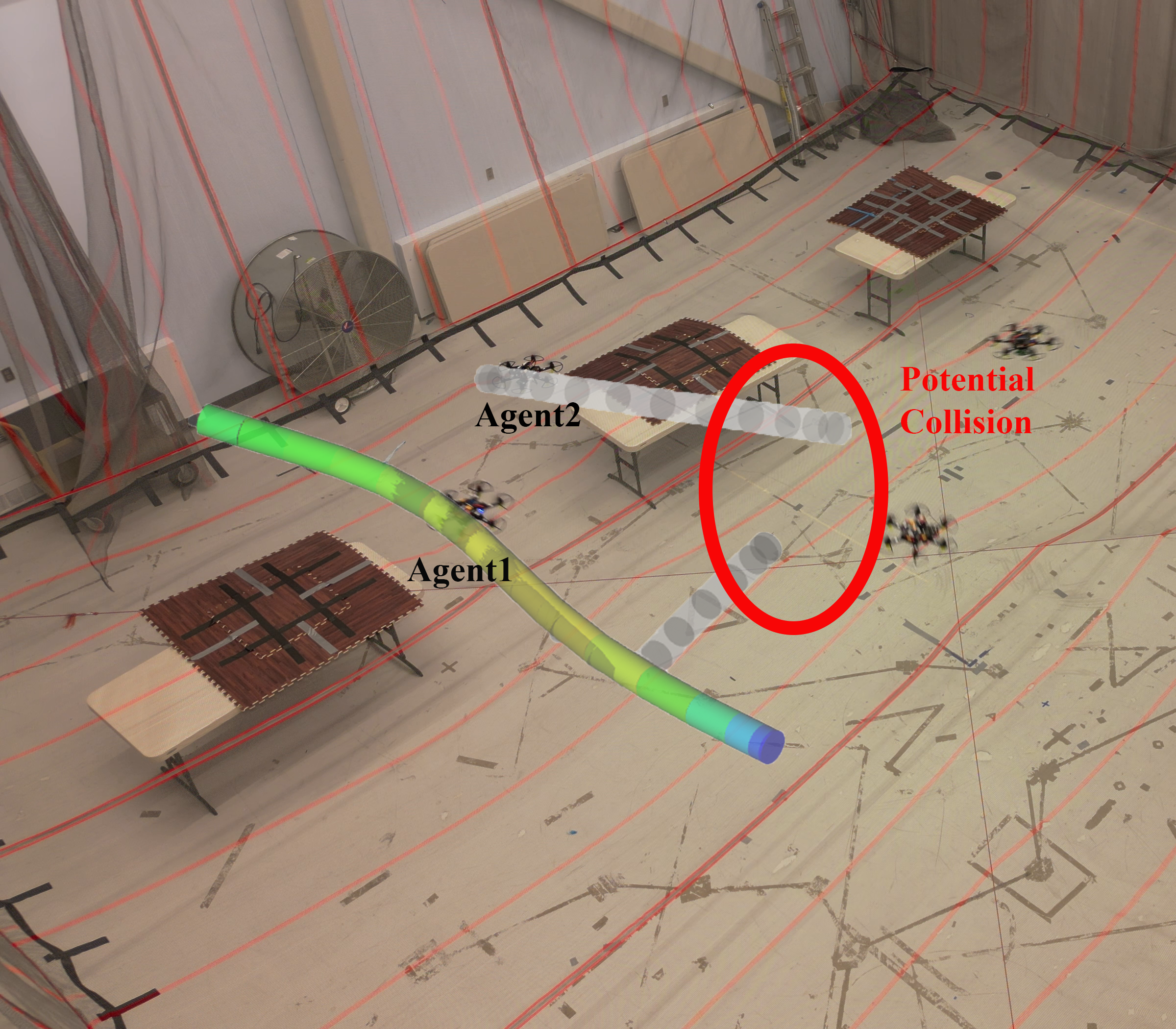}} 
    \subfloat{\includegraphics[width=0.49\columnwidth, height=0.14\textheight]{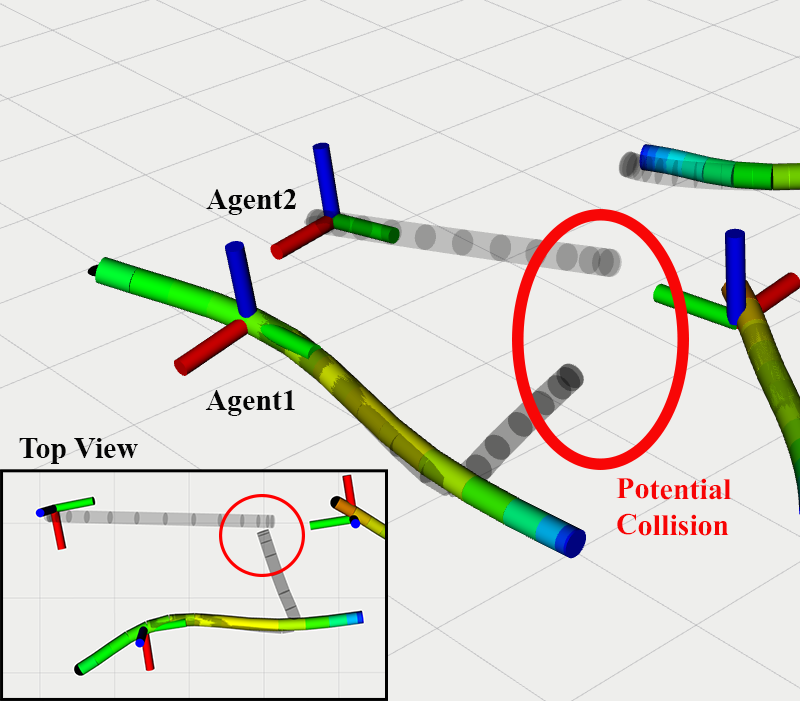}} 
    \\ {\scriptsize $t=$ \SI{0.15}{\s}: Agent 1 and Agent 2 published their traj\textsubscript{new} only \SI{10}{\ms} apart. Due to communication delays each agent did not consider the other trajectory, and thus these two trajectories are in conflicts. Note that we have a \SI{1.5}{\m}-tall boundary box, and thus these trajectories are in collision.} \\[1em]
    \subfloat{\includegraphics[width=0.49\columnwidth, height=0.14\textheight]{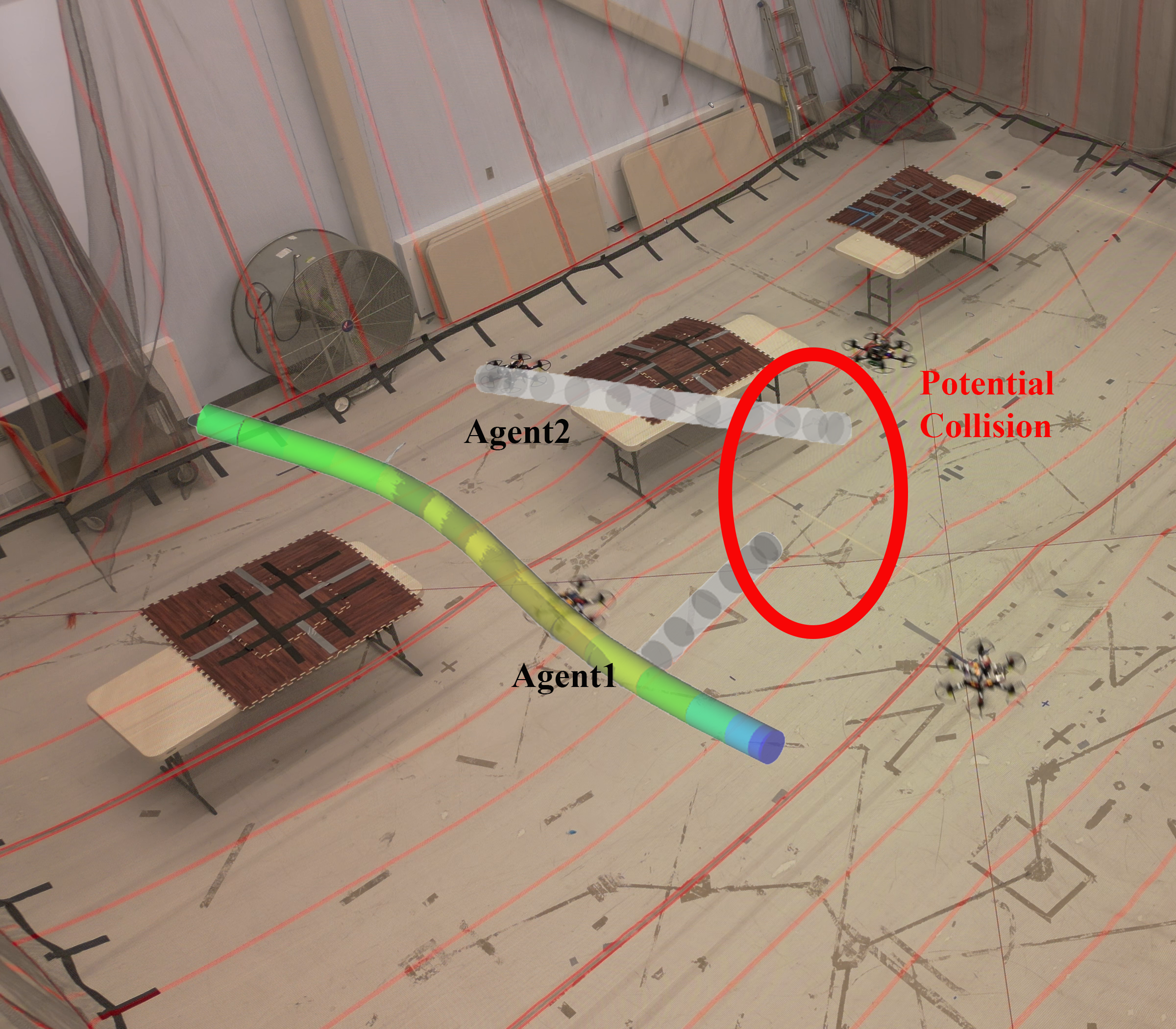}} 
    \subfloat{\includegraphics[width=0.49\columnwidth, height=0.14\textheight]{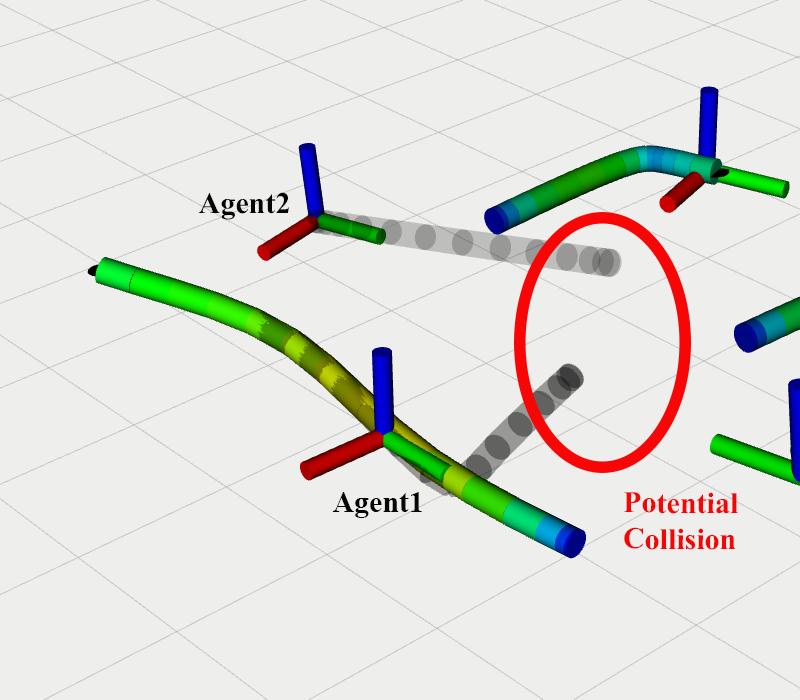}} 
    \\ {\scriptsize $t=$ \SI{1.01}{\s}: During Delay Check both agents detected conflicts and did not commit their trajectory.} \\[1em]
    \subfloat{\includegraphics[width=0.49\columnwidth, height=0.14\textheight]{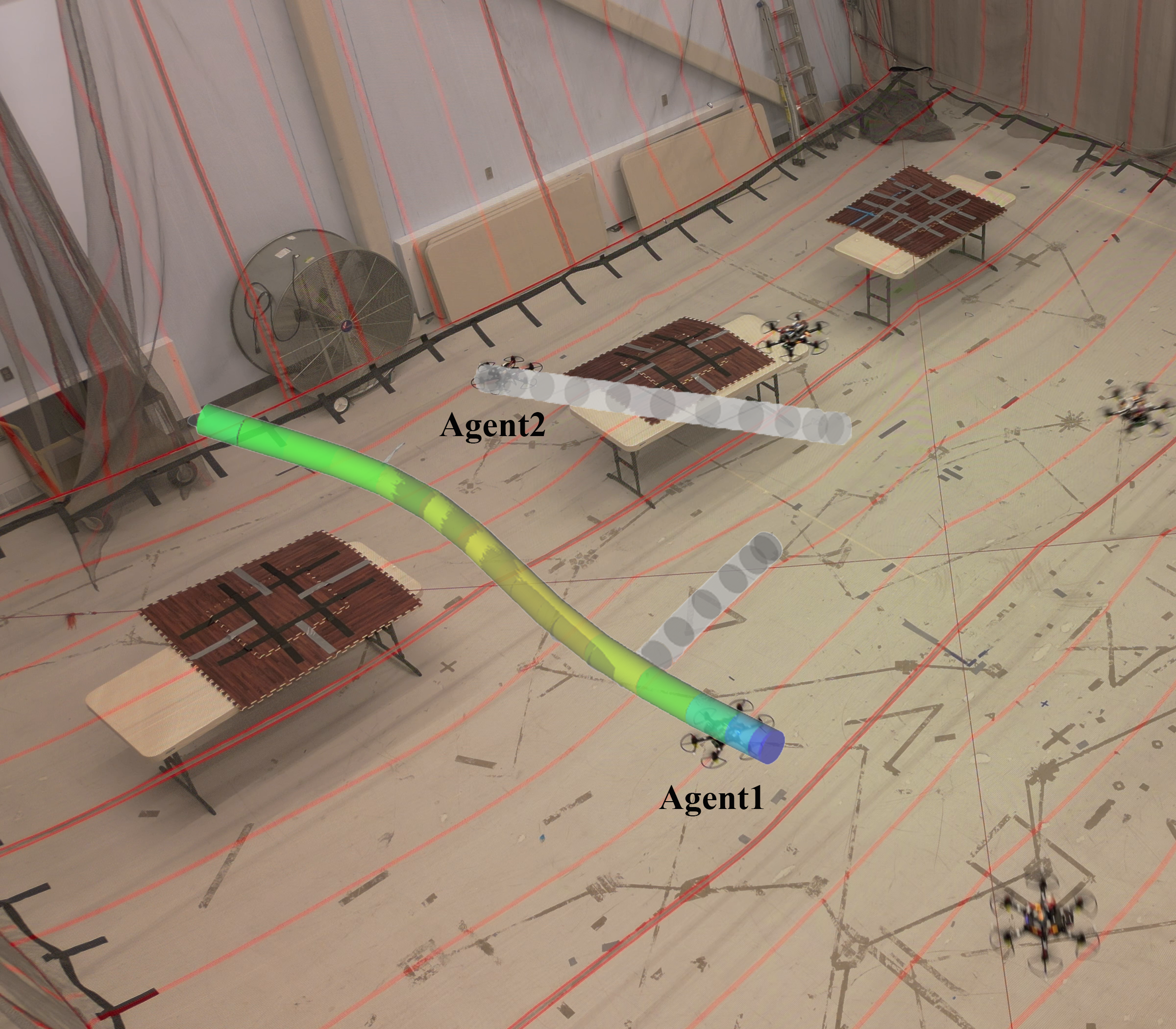}}
    \subfloat{\includegraphics[width=0.49\columnwidth, height=0.14\textheight]{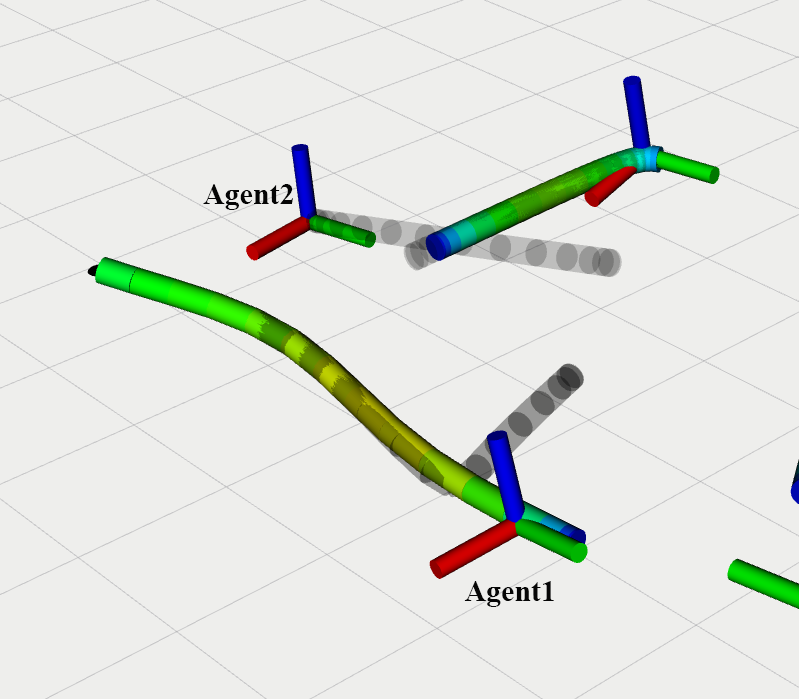}}
    \\ {\scriptsize $t=$ \SI{1.97}{\s}: Collision avoided.}
    \caption{RMADER Successful Deconfliction under Communication Delays}
    \label{fig:rmader_centr}
    \vspace{-1.5em}
\end{figure}

\ifdefined\qtyproduct
\else
  \ifdefined\NewCommandCopy
    \NewCommandCopy\qtyproduct\SI
  \else
    \NewDocumentCommand\qtyproduct{O{}mm}{\SI[#1]{#2}{#3}}
  \fi
\fi

A total of 10 hardware experiments (5 flights for each) demonstrate RMADER's robustness to communication delays as well as MADER's shortcomings. Each flight test had 6 UAVs in the \qtyproduct{9.2 x 7.5 x 2.5}{\m} flight space and lasted $\approx$\SI{1}{\minute}. The collision-safety boundary box around each UAV was set to \qtyproduct{0.8 x 0.8 x 1.5}{\m}. Note that the $z$ component of this boundary box is larger to avoid the effects of downwash from other agents. All the planning and control run onboard the UAV, and the state estimation is obtained by fusing IMU measurements with an external motion capture system. A safety mechanism running in parallel reports potential collisions and sends commands to the UAVs to avoid colliding.  

During the MADER hardware experiments due to the effects of communication delays, \textbf{7} potential collisions were detected by the safety mechanism. RMADER, on the other hand, did not generate conflicts. 
A snapshot of one of the RMADER experiments is shown in Fig.~\ref{fig:mader_hw_6_uavs}, and a successful trajectory deconfliction despite the communication delay is shown in Fig.~\ref{fig:rmader_centr}.
The maximum velocities and flight distances achieved during the RMADER hardware experiments are shown in Table~\ref{tab:max_vel_flight_dist_rmader}. The UAVs achieved maximum velocity of \SI{3.4}{\m/\s} in the third RMADER experiment.

\section{CONCLUSIONS and FUTURE WORK}

We proposed RMADER, a decentralized and asynchronous multiagent trajectory planner that is robust to communication delays. The key property of RMADER is that it guarantees safety even when there are communication delays. RMADER guarantees collision-free trajectories by introducing a delay check mechanism and keeping at least one collision-free trajectory available throughout planning. Simulation and hardware experiments showed RMADER's robustness to communication delays and the trade-off between safety and performance. Potential future work includes implementing VIO for localization and larger scale hardware experiments.




\appendices



\section{Convex vs Nonconvex MADER} \label{sec:convex_nonconvex_MADER}

\newcommand{\NextTraj}{\tikz[baseline=0.0ex]\draw [red,thick, dash pattern=on 3pt off 1pt] (0,0.08) -- (0.5,0.08);}
\newcommand{\CurrTraj}{\tikz[baseline=0.0ex]\draw [red,thick] (0,0.08) -- (0.5,0.08);}
\newcommand{\modifiedFig}[1]{\fcolorbox{white}{white}{#1}}
\newcommand{\ff}[1]{f_j^{\text{BS}\rightarrow\text{MV}}(#1)}
\newcommand{\ffQ}{\ff{\mathcal{Q}_{j}^{\text{BS}}}}
\newcommand{\ffDot}{\ff{\cdot}}
\newcommand{\hh}[1]{h_j^{\text{BS}\rightarrow\text{MV}}(#1)}
\newcommand{\hhQ}{\hh{\mathcal{Q}_{j}^{\text{BS}}}}
\newcommand{\hhDot}{\hh{\cdot}}

Our prior work MADER~\cite{tordesillas_mader_2022} formulated a nonconvex optimization problem by using both the control points and the separating planes as decision variables~\cite[Section VI-D]{tordesillas_mader_2022}. This could, however, cause expensive onboard computation. Therefore we re-formulated the problem as convex by fixing the separating planes in the optimization (i.e., by not including these planes as decision variables). In addition, to generate smoother trajectories, we added a constraint on the maximum jerk. 

We compared both version on a \texttt{general-purpose-N2} Google Cloud instance with 32 Intel\textsuperscript{\small\textregistered} Core i7. The flight space contains 250 dynamic and static obstacles, and the UAV must fly through the space to reach a goal \SI{75}{m} away. Maximum velocity/acceleration/jerk are set to \SI{10}{\m/\s} / \SI{20}{\m/\s^2} / \SI{30}{\m/\s^3}. The performance was measured in terms of \emph{Computation Time}, trajectory smoothness indicated by \intaccelsquared{} and \intjerksquared{}, \emph{Number of Stops}, \emph{Travel Time}, and \emph{Travel Distance}. The results are shown in Table~\ref{tab:mader-comparison}, where all data is the average of \textbf{100 simulations}. The notation \intaccelsquared{} and \intjerksquared{} refers to the time integral of squared norm of the acceleration and jerk along the trajectory, respectively. Higher values therefore represent a less smooth trajectory. \emph{Number of Stops} is the number of times the UAV had to stop on its way to the goal. Table~\ref{tab:mader-comparison} indicates that convex MADER is computationally less expensive and generates smoother trajectories, but nonconvex MADER performs better in terms of the \emph{Number of Stops} and \emph{Travel Time}. Since convex MADER has a computational advantage and can generate smoother trajectories, we implemented convex MADER for MADER and RMADER in all the simulations and hardware experiments in this paper.   

\begin{table}
    \centering
    \caption{\centering Convex MADER vs Nonconvex MADER \label{tab:mader-comparison}}
    \resizebox{1.0\columnwidth}{!}{%
    \begin{tabular}{>{\centering\arraybackslash}m{0.1\columnwidth} >{\centering\arraybackslash}m{0.05\columnwidth} >{\centering\arraybackslash}m{0.05\columnwidth} >{\centering\arraybackslash}m{0.13\columnwidth} >{\centering\arraybackslash}m{0.12\columnwidth} >{\centering\arraybackslash}m{0.12\columnwidth} >{\centering\arraybackslash}m{0.13\columnwidth} >{\centering\arraybackslash}m{0.18\columnwidth}}
        \toprule
        \multirow{2}{*}{Method} & \multicolumn{2}{c}{\makecell{Computation \\ Time [ms]}} & \multirow{2}{0.13\columnwidth}{\centering \intaccelsquared{} [m$^2$/s$^3$]} & \multirow{2}{0.13\columnwidth}{\centering \intjerksquared{} [m$^2$/s$^5$]} & \multirow{2}{0.12\columnwidth}{\centering Number of Stops} & \multirow{2}{0.12\columnwidth}{\centering Travel Time [s]} & \multirow{2}{0.18\columnwidth}{\centering Travel Distance [m]} \tabularnewline
        & Avg & Max & & & & & \tabularnewline
        \hline
        \makecell{convex \\ MADER} & \textbf{31.08} & \textbf{433.0} & \textbf{103.5} & \textbf{2135.0} & 0.18 & 16.05 & \textbf{75.24} \tabularnewline
        \hline
        \makecell{nonconvex \\ MADER} & 39.23 & 724.0 & 441.93 & 20201.8 &  \textbf{0.16} & \textbf{9.93} & 75.80 \tabularnewline
        \bottomrule 
    \end{tabular}}
\vspace{-2em}
\end{table}

\section*{ACKNOWLEDGMENT}
We would like to thank Nick Rober, Lakshay Sharma, Miguel Calvo-Fullana, Andrea Tagliabue, 	
Dong-Ki Kim, and Jeremy Cai, for their help, discussions, and insightful comments on this paper. This research is funded in part by Boeing Research \& Technology.

\balance

\bibliographystyle{IEEEtran}


\bibliography{bib.bib}

\end{document}